\def \figwidth {4.3cm}
\def \bigfigwidth {8.6cm}

\def \halffigwidth {4.4cm}
\def \quarterfigwidth {2.2cm}

\newif\ifsinglecol
\singlecolfalse

\documentclass[10pt,journal,letterpaper,compsoc]{IEEEtran}

\usepackage{graphicx}
\usepackage{latexsym}
\usepackage{amsfonts}
\usepackage{amsmath}
\usepackage{amsthm}
\usepackage{tocloft}
\usepackage{color}
\usepackage{multirow}

\usepackage{algorithm}
\usepackage{algpseudocode}
\algtext*{EndIf}

\usepackage{multirow}
\usepackage{url}

\ifCLASSOPTIONcompsoc
\else
\fi
\usepackage{amsbsy}

\newcommand{\T}{\mathrm{T}}

\newcommand{\diag}{\mathrm{diag}}

\newcommand{\norm}{\mathcal{N}}

\newcommand{\Ga}{\mathcal{G}}

\newcommand{\fvect}{\mathbf{f}}
\newcommand{\gvect}{\mathbf{g}}

\newcommand{\mvect}{\mathbf{m}}

\newcommand{\xvect}{\mathbf{x}}
\newcommand{\yvect}{\mathbf{y}}
\newcommand{\kvect}{\mathbf{k}}
\newcommand{\zerovect}{\mathbf{0}}

\newcommand{\thetavect}{\boldsymbol{\theta}}

\newcommand{\etavect}{\boldsymbol{\eta}}

\newcommand{\nuvect}{\boldsymbol{\nu}}
\newcommand{\muvect}{\boldsymbol{\mu}}

\usepackage[usenames,dvipsnames]{xcolor}

\begin{document}
%
\title{Pseudo-Marginal Bayesian Inference for Gaussian Processes} 
%
%
%
%

\author{Maurizio Filippone and~Mark Girolami 
\IEEEcompsocitemizethanks{\IEEEcompsocthanksitem M. Filippone is with the School of Computing Science, University of Glasgow, UK.\protect
\ E-mail: maurizio.filippone@glasgow.ac.uk 
\IEEEcompsocthanksitem M. Girolami is with the Department of Statistics, University of Warwick, UK.\protect
\ E-mail: m.girolami@warwick.ac.uk
\IEEEcompsocthanksitem First version dated 24 July 2012 was submitted to arXiv on 2 Oct 2013. Revised on 10 Apr 2013, 5 Mar 2014, and 4 Apr 2014.}
\thanks{}}

%
%

\markboth{}%
{Shell \MakeLowercase{\textit{et al.}}: Bare Demo of IEEEtran.cls for Computer Society Journals}
%



\IEEEcompsoctitleabstractindextext{%
\begin{abstract}
The main challenges that arise when adopting Gaussian Process priors in probabilistic modeling are how to carry out exact Bayesian inference and how to account for uncertainty on model parameters when making model-based predictions on out-of-sample data.
Using probit regression as an illustrative working example, this paper presents a general and effective methodology based on the pseudo-marginal approach to Markov chain Monte Carlo that efficiently addresses both of these issues.
The results presented in this paper show improvements over existing sampling methods to simulate from the posterior distribution over the parameters defining the covariance function of the Gaussian Process prior.
This is particularly important as it offers a powerful tool to carry out full Bayesian inference of Gaussian Process based hierarchic statistical models in general.
The results also demonstrate that Monte Carlo based integration of all model parameters is actually feasible in this class of models providing a superior quantification of uncertainty in predictions. 
Extensive comparisons with respect to state-of-the-art probabilistic classifiers confirm this assertion.
\end{abstract}

}

\maketitle

\IEEEdisplaynotcompsoctitleabstractindextext

%
\IEEEpeerreviewmaketitle

\section{Introduction}

Non-parametric or kernel based models represent a successful class of statistical modelling and prediction methods. To focus ideas throughout the paper we employ the working example of predictive classification problems; the methodology presented, however, is applicable to all hierarchic Bayesian models in general and those employing Gaussian Process (GP) priors in particular.
Important examples of kernel-based classifiers are the Support Vector Machine (SVM)~\cite{Cortes95,Vapnik95}, the Relevance Vector Machine (RVM)~\cite{Rasmussen05,Tipping01}, and the Gaussian Process classifier~\cite{Rasmussen06}.
Although these classifiers are based on different modeling assumptions and paradigms of statistical inference, they are characterized by a kernel function or covariance operator that allows one to build nonlinear classifiers able to tackle challenging problems~\cite{Bosch08,FilipponeAOAS12,Jaakkola00b,Joachims98,Ratsch06,OWilliams05}.

In order to allow these classifiers to be flexible, it is necessary to parameterize the kernel (or covariance) function by a set of so called {\em hyper-parameters}.
After observing a set of training data, the aim is to estimate or infer such hyper-parameters.
In the case of SVMs point estimates of hyper-parameters are obtained by optimizing a cross-validation error.
This makes optimization viable only in the case of very few hyper-parameters, as grid search is usually employed, and is limited by the available amount of data.
In GP classification, instead, the probabilistic nature of the model provides a means (usually after approximately integrating out latent variables) to obtain an approximate marginal likelihood 
that offers the possibility to optimize the hyper-parameters; this is known as type II Maximum Likelihood (ML)~\cite{Bishop06,Rasmussen06}.
Deterministic approximations for integrating out the latent variables include the Laplace Approximation (LA)~\cite{Williams98}, Expectation Propagation (EP)~\cite{Minka01}, Variational Bayes~\cite{Gibbs00}, Integrated Nested Laplace Approximations~\cite{Rue09}, and mean field approximations~\cite{Opper00}; see~\cite{Kuss05,Nickisch08} for extensive assessments of the relative merits of different approximation schemes for GP classification.

From a fully Bayesian perspective, in a GP classifier one would like to be able to (i) infer all model parameters and latent variables from data and (ii) integrate out latent variables and hyper-parameters with respect to their posterior distribution when making predictions accounting for their uncertainty; this in particular, would effectively make the classifier {\em parameter free}.
To date, the literature lacks a systematic way to efficiently tackle both of these questions.
The main limitations are due the fact that it is not possible to obtain any of the quantities needed in the inference in closed form because of analytical intractability.
This requires the use of some form of approximation, and deterministic approximations have been proposed to integrate out latent variables only; in order to integrate out the hyper-parameters most of the approaches propose quadrature methods~\cite{Cseke11,Rue09}, that can only be employed in the case of a small number of hyper-parameters.

Recently, there have been a few attempts to carry out inference using stochastic approximations based on Markov chain Monte Carlo (MCMC) methods~\cite{FilipponeML13,Murray10,KnorrHeld02}, the idea being to leverage asymptotic guarantees of convergence of Monte Carlo estimates to the true values.
Unfortunately, employing MCMC methods for inferring latent variables and hyper-parameters is extremely challenging, and state-of-the-art methods for doing so are still inefficient and difficult to use in practice. 

This paper aims at providing a straightforward to implement methodology that is effective in the direction of bridging this gap, by proposing an MCMC method that addresses most of the difficulties that one is faced with when applying stochastic based inference in GP modeling, such as discrete label classification.
The main issue in applying MCMC methods to carry out inference in GP classification is in sampling the hyper-parameters form the full posterior.
This is due to the structure of the model that makes latent variables and hyper-parameters strongly coupled a posteriori; as a result, chains are characterized by low efficiency and poor mixing.
The key idea of the proposed methodology is to break the correlation in the sampling between latent variables and hyper-parameters by approximately integrating out the latent variables while retaining a correct MCMC procedure; namely, maintaining the exact posterior distribution over hyper-parameters as the invariant distribution of the chains and ergodicity properties.
This can be achieved by means of the so called {\em Pseudo Marginal} (PM) approach to Monte Carlo sampling~\cite{Andrieu09,Beaumont03}.
This work shows that the use of the PM approach leads to remarkable efficiency in the sampling of hyper-parameters, thus making the fully Bayesian treatment viable and simple to implement and employ.

The importance of integrating out the hyper-parameters to achieve a sound quantification of uncertainty in predictions is well known and has been highlighted, for example, in~\cite{Bishop06,Neal99,Rue09,Taylor12}; employing this in practice, however, is notoriously challenging.
The main motivation for this work, is to demonstrate that this marginalization can be done exactly, in a Monte Carlo sense, by building upon deterministic approximations already proposed in the GP and Machine Learning literature.
This work reports a thorough empirical comparison in this direction, showing the ability of the fully Bayesian treatment to achieve a better quantification of uncertainty compared to the standard practice of optimization of the hyper-parameters in GP classification. 
Furthermore, the results report a comparison with a probabilistic version of the SVM classifier~\cite{Chang11}.
The results on GP classification support the argument that hyper-parameters should be integrated out to achieve a reliable quantification of uncertainty in applications and this paper provides a practical means to achieve this\footnote{The code to reproduce all the experiments is available at: \\\url{http://www.dcs.gla.ac.uk/~maurizio/pages/code_pm/}}. \label{REV:2:2}

The paper is organized as follows: 
section~2 reviews the Gaussian Process approach to classification, and section~3 presents the proposed MCMC approach to obtain samples from the posterior distribution over both latent variables and hyper-parameters.
Section~4 reports an assessment of the sampling efficiency achieved by the PM approach compared to other MCMC approaches, and section~5 reports a study on the performance of the fully Bayesian GP classifier compared to other probabilistic classifiers.
Finally, section~6 concludes the paper.

\section{Gaussian Process Classification}

In this section, we briefly review GP classification based on a probit likelihood (see~\cite{Rasmussen06} for an extensive presentation of GPs). 
Let $X = \{\xvect_1, \ldots, \xvect_n \}$ be a set of $n$ input vectors described by $d$ covariates and associated with observed univariate responses $\yvect = \{y_1, \ldots, y_n\}$ with $y_i \in \{-1, +1\}$.
Let $\fvect = \{f_1, \ldots, f_n \}$ be a set of latent variables.
From a generative perspective, GP classifiers assume that the class labels have a Bernoulli distribution with success probability given by a transformation of the latent variables:
\begin{equation}
p(y_i | f_i) = 
\Phi(y_i f_i).
\end{equation}
Here $\Phi$ denotes the cumulative function of the Gaussian density; based on this modeling assumption, the likelihood function is: \label{REV:1:13}
\begin{equation}
p(\yvect | \fvect) = \prod_{i=1}^n p(y_i | f_i).
\end{equation}
The latent variables $\fvect$ are given a zero mean GP prior with covariance $K$:
\begin{equation}
\fvect \sim \norm(\fvect | \zerovect, K).
\end{equation}

Let $k(\xvect_i, \xvect_j | \thetavect)$ be the function modeling the covariance between latent variables evaluated at the input vectors, parameterized by a vector of hyper-parameters $\thetavect$.
In this paper we will adopt a covariance function defined as follows: \label{REV:2:14}
\begin{equation}
k(\xvect_i, \xvect_j | \thetavect) = 
\sigma \exp\left[-\frac{1}{2} (\xvect_i - \xvect_j)^{\T} A (\xvect_i - \xvect_j)  \right].
\end{equation}
The parameter $\sigma$ is the variance of the marginal prior distribution for each of the latent variables $f_i$.  \label{REV:2:3}
The matrix $A$, instead, defines the type of covariance between the values of the function at different input vectors.
By defining a matrix $A$ with a global parameter as follows,
\begin{equation} \label{eq:covariance:isotropic}
A^{-1} = \tau^2 I,
\end{equation}
an isotropic covariance function is obtained.
Alternatively, $A$ can be defined to assign a different parameter to each covariate
\begin{equation}
A^{-1} = \diag\left(\tau^2_1, \ldots, \tau^2_d\right).
\end{equation}
The latter choice yields the so called {\em Automatic Relevance Determination} (ARD) prior~\cite{Mackay94}.
In this formulation, the hyper-parameters $\tau_i$ can be interpreted as length-scale parameters.
Let $\thetavect$ be a vector comprising $\sigma$ and all the length-scale parameters, and $K$ be the matrix whose entries are $k_{ij} = k(\xvect_i, \xvect_j |  \thetavect)$.

The GP classification model is hierarchical, as $\yvect$ is conditioned on $\fvect$, and $\fvect$ is conditioned on $\thetavect$ and the inputs $X$.
In order to keep the notation uncluttered, in the remainder of this paper we will not report explicitly the conditioning on the inputs in any of the equations.
We now briefly review the types of approximations that have been proposed in the literature to employ GP classifiers.

\subsection{Deterministic approximations for integrating out latent variables}

One of the difficulties encountered in GP classification is that, unlike GP regression, the prior on the latent variables and the likelihood do not form a conjugate pair; therefore, it is not possible to analytically integrate out the latent variables.
As a consequence, it is not possible to directly sample from or optimize the distribution of hyper-parameters given the labels, nor directly evaluate predictive probabilities.
This has motivated a large body of research that attempts to approximate the posterior distribution over the latent variables $p(\fvect | \yvect, \thetavect)$ with a Gaussian $q(\fvect | \yvect, \thetavect) = \norm(\fvect | \mu_q, \Sigma_q)$ in order to exploit conjugacy.
By doing so, it is possible to analytically integrate out latent variables to obtain an approximate marginal likelihood, and compute the predictive distribution for new data, as discussed in the following.
The Gaussian approximation yields an approximate marginal likelihood $\hat{p}(\yvect | \thetavect)$ that can then be optimized with respect to the hyper-parameters, or used to obtain samples from the approximate posterior distribution over the hyper-parameters, say $\hat{p}(\thetavect | \yvect)$, using MCMC techniques.
We now briefly discuss how this can be achieved.

To obtain an approximate predictive distribution, conditioned on a value of the hyper-parameters $\thetavect$, we can compute:
\begin{equation} \label{eq:approx:integration:f}
p(y_* | \yvect, \thetavect) = \int p(y_* | f_*) p(f_* | \fvect, \thetavect) q(\fvect | \yvect, \thetavect) df_* d\fvect.
\end{equation}
Here $\thetavect$ can be a ML estimate that maximizes the approximate likelihood or one sample from the approximate posterior $\hat{p}(\thetavect | \yvect)$.
For simplicity of notation, let $K$ be the covariance matrix evaluated at $\thetavect$, $\kvect_*$ the vector whose $i$th element is $k(\xvect_i, \xvect_* | \thetavect)$ and $k_{**} = k(\xvect_*, \xvect_* | \thetavect)$.
Given the properties of multivariate normal variables, $f_*$ is distributed as $\norm(f_* | \mu_*, \beta^2_*)$ with $\mu_* = \kvect_*^{\T} K^{-1} \fvect$ and $\beta^2_* = k_{**} - \kvect_*^{\T} K^{-1} \kvect_*$. \label{REV:2:15}
Approximating $p(\fvect | \yvect, \thetavect)$ with a Gaussian $q(\fvect | \yvect, \thetavect) = \norm(\fvect | \muvect_q, \Sigma_q)$ makes it possible to analytically perform integration with respect to $\fvect$ in equation \ref{eq:approx:integration:f}.
In particular, the integration with respect to $\fvect$ yields $\norm(f_* | m_*, s^2_*)$ with
$$
m_* = \kvect_*^{\T} K^{-1} \muvect_{q},
$$
and 
$$
s^2_* = k_{**} - \kvect_*^{\T} K^{-1} \kvect_* + \kvect_*^{\T} K^{-1} \Sigma_{q} K^{-1} \kvect_*.
$$
The univariate integration with respect to $f_*$ follows exactly in the case of a probit likelihood, as it is a convolution of a Gaussian and a cumulative Gaussian 
\begin{equation}
\int p(y_* | f_*) \norm(f_* | m_*, s^2_*) df_* = \Phi\left( \frac{m_*}{\sqrt{1 + s^2_*}} \right).
\end{equation}
We now briefly review two popular approximation methods for integrating out latent variables, namely the Laplace Approximation and Expectation Propagation. \label{REV:2:5}

\subsubsection{Laplace Approximation}
The Laplace Approximation (LA) is based on the assumption that the distribution of interest can be approximated by a Gaussian centered at its mode and with the same curvature.
By analyzing the Taylor expansion of the logarithm of target and approximating densities, the latter requirement is satisfied by imposing an inverse covariance for the approximating Gaussian equal to the negative Hessian of the logarithm of the target density~\cite{Tierney86}.
For a given value of the hyper-parameters $\thetavect$, define
\begin{equation} \label{eq:laplace:target}
\Psi(\fvect) = \log[p(\yvect | \fvect)] + \log[p(\fvect | \thetavect)] + \mathrm{const.}
\end{equation}
as the logarithm of the target density up to terms independent of $\fvect$.
Performing a Laplace approximation amounts in defining a Gaussian $q(\fvect | \yvect, \thetavect) = \norm(\fvect | \hat{\fvect}, \hat{\Sigma})$, such that
\begin{equation} \label{eq:fhat}
\hat{\fvect} = \arg\max_{\fvect} \Psi(\fvect) \qquad \mathrm{and} \qquad
\hat{\Sigma}^{-1} = - \nabla_{\fvect} \nabla_{\fvect} \Psi(\hat{\fvect}).
\end{equation}
As it is not possible to directly solve the maximization problem in equation~\ref{eq:fhat}, an iterative procedure based on the following Newton-Raphson formula is usually employed:
\begin{equation} \label{eq:laplace:newton:iteration}
\fvect_{\mathrm{new}} = \fvect - (\nabla_{\fvect} \nabla_{\fvect} \Psi(\fvect))^{-1} \nabla_{\fvect} \Psi(\fvect)
\end{equation}
starting from $\fvect = \zerovect$ until convergence.
The gradient and the Hessian of the log of the target density are:
\begin{equation}
\nabla_{\fvect} \Psi(\fvect) = \nabla_{\fvect} \log[p(\yvect | \fvect)] - K^{-1} \fvect,
\end{equation}
\begin{equation}
\nabla_{\fvect} \nabla_{\fvect} \Psi(\fvect) = \nabla_{\fvect} \nabla_{\fvect} \log[p(\yvect | \fvect)] - K^{-1}. 
\end{equation}
Note that if $\log[p(\yvect | \fvect)]$ is concave, such as in probit classification, $\Psi(\fvect)$ has a unique maximum.
Practically, the Newton-Raphson update in equation \ref{eq:laplace:newton:iteration} is implemented by employing Woodbury identities to avoid inverting $K$ directly (see section 3.4 of~\cite{Rasmussen06} for full details).
In such an implementation, one $n \times n$ matrix factorization is needed at each iteration and no other $O(n^3)$ operations. 

\subsubsection{Expectation Propagation}
The Expectation Propagation (EP) algorithm is based on the assumption that each individual term of the likelihood can be approximated by an unnormalized Gaussian
\begin{equation}
p(y_i | f_i) \simeq \tilde{Z}_i \norm(f_i | \tilde{\mu}_i, \tilde{\sigma}^2_i).
\end{equation}
Approximating each term in the likelihood by a Gaussian implies that the approximate likelihood, as a function of $\fvect$, is multivariate Gaussian
\begin{equation}
\norm(\fvect | \tilde{\muvect}, \tilde{\Sigma}) \prod_{i=1}^n \tilde{Z}_i
\end{equation}
with $\tilde{\muvect}_i = \tilde{\mu}_i$ and $\tilde{\Sigma}_{ii} = \tilde{\sigma}_i$.

Under this approximation, the posterior $p(\fvect | \yvect, \thetavect)$ is approximated by a Gaussian $q(\fvect | \yvect, \thetavect) = \norm(\fvect | \hat{\fvect}, \hat{\Sigma})$ with:
\begin{equation}
\hat{\Sigma}^{-1} = K^{-1} + \tilde{\Sigma}^{-1}  \qquad \mathrm{and} \qquad
\hat{\fvect} = \hat{\Sigma} \, \tilde{\Sigma} \, \tilde{\muvect}
\end{equation}

The EP algorithm is characterized by the way the parameters $\tilde{Z}_i$, $\tilde{\mu}_i$, and $\tilde{\sigma}^2_i$ are optimized.
The EP algorithm loops through the $n$ factors approximating the likelihood updating those three parameters for each factor in turn.
First, the so called {\em cavity distribution} is computed
\begin{equation}
q^{\setminus i}(f_i | \thetavect) \propto \int p(\fvect | \thetavect) \prod_{j \neq i}  \tilde{Z}_j \norm(f_j | \tilde{\mu}_j, \tilde{\sigma}^2_j), 
\end{equation}
which is obtained by leaving out the $i$th factor from $q(\fvect | \yvect, \thetavect)$.
Second, a revised Gaussian $q^{\prime}(f_i | \thetavect)$, which closely approximates the product of the cavity distribution and the exact $i$th likelihood term, is sought.
In particular, this is performed by minimizing the following Kullback-Leibler divergence:
\begin{equation}
\mathrm{KL}\left(
\left. q^{\setminus i}(f_i | \thetavect) p(y_i | f_i) \right\| 
q^{\prime}(f_i | \thetavect)
\right),
\end{equation}
which in practice boils down to matching the moments of the two distributions.
Third, once the mean and variance of $q^{\prime}(f_i | \thetavect)$ are computed, it is possible to derive the updated parameters $\tilde{Z}_i$, $\tilde{\mu}_i$, and $\tilde{\sigma}^2_i$ for the $i$th factor.
The derivation of those equations is rather involved, and the reader is referred to~\cite{Rasmussen06} for full details; EP requires five operations in $O(n^3)$ at each iteration.
Note that convergence of the EP algorithm is not guaranteed in general; however, for GP classification, no convergence issues have been reported in the literature.
Furthermore, EP for GP classification has been reported to offer superior accuracy in approximations compared to other methods~\cite{Kuss05,Nickisch08}.

\subsection{Fully Bayesian treatment}

In a fully Bayesian treatment, the aim is to integrate out latent variables as well as hyper-parameters:
\begin{equation}
p(y_* | \yvect) = \int p(y_* | f_*) p(f_* | \fvect, \thetavect) p(\fvect, \thetavect | \yvect) df_* d\fvect d\thetavect.
\end{equation}
Again, the integration with respect to $f_*$ can be done analytically, whereas the integration with respect to latent variables and hyper-parameters requires the posterior distribution $p(\fvect, \thetavect | \yvect)$.
One way to tackle the intractability in characterizing $p(\fvect, \thetavect | \yvect)$ is to draw samples from $p(\fvect, \thetavect | \yvect)$ using MCMC methods, so that a Monte Carlo estimate of the predictive distribution can be used
\begin{equation} \label{eq:monte:carlo:integration}
p(y_* | \yvect) \simeq \frac{1}{N} \sum_{i=1}^N \int p(y_* | f_*) p(f_* | \fvect^{(i)}, \thetavect^{(i)}) df_* ,
\end{equation}
where $\fvect^{(i)}, \thetavect^{(i)}$ denotes the $i$th sample from $p(\fvect, \thetavect | \yvect)$.
This estimate will asymptotically converge to the exact expectation $p(y_* | \yvect)$.

\section{MCMC Sampling From $p(\fvect, \thetavect | \yvect)$}

Sampling from the posterior over $\fvect$ and $\thetavect$ by joint proposals is not feasible; it is extremely unlikely to propose a set of latent variables and hyper-parameters that are compatible with each other and observed data.
In order to draw samples from $p(\fvect, \thetavect | \yvect)$, it is therefore necessary to resort to a Gibbs sampler, whereby $\fvect$ and $\thetavect$ are updated in turn.
We now briefly review the state of the art in Gibbs sampling techniques for GP models, and propose a new Gibbs sampler based on the PM approach.

\subsection{Drawing samples from $p(\fvect | \yvect, \thetavect)$}
Efficient sampling of the latent variables can be achieved by means of Elliptical Slice Sampling (ELL-SS)~\cite{Murray10b}.
ELL-SS is based on an adaptation of the Slice Sampling algorithm~\cite{Neal03} to propose new values of the latent variables.
ELL-SS has the very appealing property of requiring no tuning, so that minimum user intervention is needed, and by the fact that once $K$ is factorized the complexity of iterating ELL-SS is in $O(n^2)$.
Recently, the efficiency of ELL-SS has been extensively demonstrated on several models involving GP priors~\cite{FilipponeML13}.

Another way to efficiently sample latent variables in GP models is by means of a variant of Hybrid Monte Carlo (HMC)~\cite{Duane87,Neal93} where the inverse mass matrix is set to the GP covariance $K$, as described in detail in~\cite{FilipponeML13}.
This variant of HMC can be interpreted as a simplified version of Riemann manifold Hamiltonian Monte Carlo (RMHMC)~\cite{Girolami11} which makes it possible to obtain samples from the posterior distribution over $\fvect$ in $O(n^2)$ once $K$ is factorized.
Owing to its simplicity, in the remainder of this paper we will use ELL-SS to sample from the posterior over latent variables $p(\fvect | \yvect, \thetavect)$.

\subsection{Drawing samples from the posterior over $\thetavect$ employing reparameterization techniques}

\subsubsection{SA and AA parameterizations}
In GP classification, efficiently sampling from the posterior distribution over the latent variables and hyper-parameters is complex because of their strong coupling~\cite{FilipponeML13,Murray10,Neal99}.
The result of this strong coupling is that fixing $\fvect$ induces a sharply peaked posterior over $\thetavect$ that makes the chain converge slowly and mix very poorly.
This effect is illustrated in Figure~\ref{fig:compare:parameterizations}. 
In particular, conditioning the sampling of $\thetavect$ on $\fvect$ corresponds to considering the standard parameterization of GP models $\yvect | \fvect$ and $\fvect | \thetavect$, which is also known as Sufficient Augmentation (SA)~\cite{Yu11}.

A better parameterization can be given by introducing a set of transformed (whitened) latent variables $\nuvect$~\cite{Papaspiliopoulos07}. 
The way $\nuvect$ is defined is by $\fvect = L \nuvect$, $L$ being the Cholesky factor of $K$.
In this parameterization, that is also known as Ancillary Augmentation (AA)~\cite{Yu11}, $\nuvect$ are constructed to be a priori independent from the hyper-parameters (using $L$ is convenient as it is needed also to evaluate the GP prior density).
In the AA parameterization $\thetavect$ is sampled from $p(\thetavect | \yvect, \nuvect)$.
The effect of conditioning on $\nuvect$ makes the conditional posterior over $\thetavect$ larger, as illustrated in Figure~\ref{fig:compare:parameterizations}. 

\subsubsection{The Surrogate data model}

In the Surrogate (SURR) data model proposed in~\cite{Murray10}, a set of auxiliary latent variables $\gvect$ is introduced as a noisy version of $\fvect$; in particular, $p(\gvect | \fvect, \thetavect) = \norm(\gvect | \fvect, S_{\thetavect})$.
This construction yields a conditional for $\fvect$ of the form $p(\fvect | \gvect, \thetavect) = \norm(\fvect | \mvect, R)$, with $R = S_{\thetavect} - S_{\thetavect} (S_{\thetavect} + K)^{-1} S_{\thetavect}$ and $\mvect = R S_{\thetavect}^{-1} \gvect$.
After decomposing $R = D D^{\T}$, the sampling of $\thetavect$ is then conditioned on the ``whitened'' variables $\etavect$, defined as $\fvect = D \etavect + \mvect$.
The covariance $S_{\thetavect}$ is constructed by matching the posterior distribution over each of the latent variables individually (see~\cite{Murray10} for further details).
Figure~\ref{fig:compare:parameterizations} shows that the SURR parameterization is characterized by a conditional posterior over $\thetavect$ larger than SA and slightly larger than the AA parameterization.


\subsection{Drawing samples from $p(\thetavect | \yvect)$: the Pseudo Marginal approach}


The use of reparameterization techniques mitigates the problems due to the coupling of latent variables and hyper-parameters, but sampling efficiency for GP models is still an issue (for example,~\cite{FilipponeAOAS12} reports simulations of ten parallel chains comprising five millions samples each).
Intuitively, the best strategy to break the correlation between latent variables and hyper-parameters in sampling from the posterior over the hyper-parameters would be to integrate out the latent variables altogether.
As we discussed, this is not possible, but here we present a strategy that uses an unbiased estimate of the marginal likelihood $p(\yvect | \thetavect)$ to devise an MCMC strategy that produces samples from the correct posterior distribution $p(\thetavect | \yvect)$.
For the sake of clarity, in this work we will focus on the Metropolis-Hastings algorithm with proposal $\pi(\thetavect^{\prime} | \thetavect)$. 
We are interested in sampling from the posterior distribution
\begin{equation}
p(\thetavect | \yvect) \propto p(\yvect | \thetavect) p(\thetavect).
\end{equation}
In order to do that, we would need to integrate out the latent variables:
\begin{equation}
p(\yvect | \thetavect) = \int p(\yvect | \fvect) p(\fvect | \thetavect) d\fvect
\end{equation}
and use this along with the prior $p(\thetavect)$ in the Hastings ratio:
\begin{equation}
z = \frac{p(\yvect | \thetavect^{\prime}) p(\thetavect^{\prime})}{p(\yvect | \thetavect) p(\thetavect)} \frac{\pi(\thetavect | \thetavect^{\prime})}{\pi(\thetavect^{\prime} | \thetavect)}
\end{equation}
As already discussed, analytically integrating out $\fvect$ is not possible.

The results in~\cite{Beaumont03,Andrieu09} show that we can plug into the Hastings ratio an estimate $\tilde{p}(\yvect | \thetavect)$ of the marginal $p(\yvect | \thetavect)$, and as long as this is unbiased, then the sampler will draw samples from the correct posterior $p(\thetavect | \yvect)$.
\begin{equation}
\tilde{z} = \frac{\tilde{p}(\yvect | \thetavect^{\prime}) p(\thetavect^{\prime})}{\tilde{p}(\yvect | \thetavect) p(\thetavect)} \frac{\pi(\thetavect | \thetavect^{\prime})}{\pi(\thetavect^{\prime} | \thetavect)}
\end{equation}
This result is remarkable as it gives a simple recipe to be used in hierarchical models to tackle the problem of strong coupling between groups of variables when using MCMC algorithms.

\begin{figure}[ht]
  \begin{center}
     \includegraphics[width=\bigfigwidth,height=\figwidth]{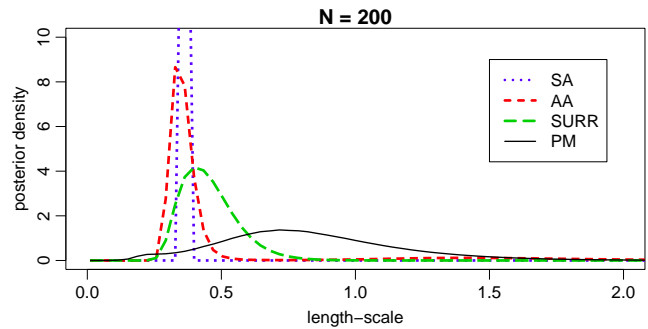}
    \caption{Comparison of the posterior distribution $p(\thetavect | \yvect)$ with the posterior $p(\thetavect | \fvect)$ in the SA parameterization, the posterior $p(\thetavect | \yvect, \nuvect)$ in the AA parameterization, and the parameterization used in the SURR method.}  \label{fig:compare:parameterizations}
  \end{center}
\end{figure}

Figure~\ref{fig:compare:parameterizations} shows the effect of conditioning the sampling of $\thetavect$ on different transformations of the latent variables given by SA (blue line), AA (red line), and SURR (green line).
The conditional variance for the three approaches is still way lower than the variance of the marginal posterior $p(\thetavect | \yvect)$ that can be obtained by the PM approach.
This motivates the use of the PM approach to effectively break the correlation between latent variables and hyper-parameters in an MCMC scheme.

Note that if the goal is quantifying uncertainty in the parameters only, and no predictions are needed, one could just iterate the sampling of $\thetavect | \yvect$, as this is done regardless of $\fvect$.   \label{REV:1:2}
For predictions, instead, samples from the joint posterior $p(\fvect, \thetavect | \yvect)$ are needed in the Monte Carlo integral in equation \ref{eq:monte:carlo:integration}, so both steps are necessary.
We consider this as a Gibbs sampler despite the fact that in principle interleaving of the two steps is not needed; one could obtain samples from the posterior distribution over $\fvect$ in a second stage, once samples from $p(\thetavect | \yvect)$ are available.
This would come at an extra cost given that sampling $\fvect | \yvect, \thetavect$ requires the factorization of $K$ for each MCMC sample $\thetavect$.
Therefore, when predictions are needed, we prefer to interleave the two steps, and still interpret the proposed sampling strategy as a Gibbs sampler.

\subsubsection{Unbiased estimation of $p(\yvect | \thetavect)$ using importance sampling}

In order to obtain an unbiased estimator $\tilde{p}(\yvect | \thetavect)$ for the marginal $p(\yvect | \thetavect)$, we propose to employ importance sampling.
We draw $N_{\mathrm{imp}}$ samples $\fvect_i$ from the approximating distribution $q(\fvect | \yvect, \thetavect)$, so that we can approximate the marginal $p(\yvect | \thetavect) = \int p(\yvect | \fvect) p(\fvect | \thetavect) d\fvect$ by:
\begin{equation} \label{eq:importance:pseudo}
\tilde{p}(\yvect | \thetavect) \simeq  \frac{1}{N_{\mathrm{imp}}} \sum_{i=1}^{N_{\mathrm{imp}}} \frac{p(\yvect | \fvect_i) p(\fvect_i | \thetavect)}{q(\fvect_i | \yvect, \thetavect)}
\end{equation}
It is easy to verify that equation~\ref{eq:importance:pseudo} yields an unbiased estimate of $p(\yvect | \thetavect)$, as its expectation is the exact marginal $p(\yvect | \thetavect)$.
Therefore, this estimate can be used in the Hastings ratio to construct an MCMC approach that samples from the correct invariant distribution $p(\thetavect | \yvect)$.
Algorithm~1 sketches the MH algorithm that we propose to sample the hyper-parameters. \label{REV:2:10} \label{REV:AE:3}
\begin{algorithm}[h!] \label{alg:MH} 
\caption{Pseudo-marginal MH transition operator to sample $\thetavect$.}
\textbf{Input}: The current pair $(\theta, \tilde{p}(\yvect | \thetavect))$, a routine to approximate $p(\fvect | \yvect, \thetavect)$ by $q(\fvect | \yvect, \thetavect)$, and number of importance samples $N_{\mathrm{imp}}$ \\
\textbf{Output}: A new pair $(\theta, \tilde{p}(\yvect | \thetavect))$
\begin{algorithmic}[1]
\State Draw $\thetavect^{\prime}$ from the proposal distribution $\pi(\thetavect^{\prime} | \thetavect)$ 
\State Approximate $p(\fvect | \yvect, \thetavect^{\prime})$ by $q(\fvect | \yvect, \thetavect^{\prime})$
\State Draw $N_{\mathrm{imp}}$ samples from $q(\fvect | \yvect, \thetavect^{\prime})$
\State Compute $\tilde{p}(\yvect | \thetavect^{\prime})$ using eq. \ref{eq:importance:pseudo}
\State Compute $\displaystyle A = \min \left\{1, \frac{\tilde{p}(\yvect | \thetavect^{\prime}) p(\thetavect^{\prime})}{\tilde{p}(\yvect | \thetavect) p(\thetavect)} \frac{\pi(\thetavect | \thetavect^{\prime})}{\pi(\thetavect^{\prime} | \thetavect)} \right\}$
\State Draw $u$ from $U_{[0,1]}$
\If {$A > u$} 
\Return $(\thetavect^{\prime}, \tilde{p}(\yvect | \thetavect^{\prime}))$
\Else {}
\Return $(\thetavect, \tilde{p}(\yvect | \thetavect))$
\EndIf
\end{algorithmic}
\end{algorithm}

From the theory of importance sampling~\cite{Robert05}, the variance of the estimator is zero when $q(\fvect | \yvect, \thetavect)$ is proportional to $p(\yvect | \fvect) p(\fvect | \thetavect)$, which is proportional to the posterior distribution over $\fvect$ that we do not know how to sample from in the first place.
In our case, the more accurate the Gaussian approximation the smaller the variance of the estimator.
Given that EP has been reported to be more accurate in approximating $p(\fvect | \yvect, \thetavect)$, it is reasonable to expect that EP will lead to a smaller estimator variance compared to LA. 
This will be assessed in the next section.

The motivation for using an importance sampling estimator rather than other simulation based methods for estimating marginal likelihoods, is the following.
Even though it is possible to sample $\fvect$ relatively efficiently, the estimation of marginal likelihoods from MCMC simulations is generally challenging~\cite{Friel08,Skilling06} and only guarantees of estimator consistency are available.
Obtaining estimates based on samples from $p(\fvect | \yvect, \thetavect)$ would require some form of user intervention (assessment of convergence and estimation of efficiency)  every time a new value of $\thetavect$ is proposed; this is clearly not practical or useful for the PM scheme.
This work reports an extensive assessment of LA and EP to obtain Gaussian approximations to $p(\fvect | \yvect, \thetavect)$ within the importance sampling estimate of $p(\yvect | \thetavect)$.  \label{REV:2:6}

\subsubsection{Analysis of correctness}
We show here why the proposed method yields an MCMC approach that produces samples from the correct invariant distribution $p(\thetavect | \yvect)$. \label{REV:1:1} \label{REV:AE:1} \label{REV:2:4}
The easiest way to see this is by considering $N_{\mathrm{imp}} = 1$; showing correctness for larger numbers of importance samples is similar but notationally heavier (see \cite{Beaumont03} for further details).
By substituting the importance sampling estimate $\tilde{p}(\yvect | \thetavect)$ with $N_{\mathrm{imp}} = 1$ into $\tilde{z}$ and rearranging the terms, we obtain
\begin{equation}
\tilde{z} = 
\frac{p(\yvect | \fvect^{\prime}) p(\fvect^{\prime} | \thetavect^{\prime}) p(\thetavect^{\prime})}{p(\yvect | \fvect) p(\fvect | \thetavect) p(\thetavect)}
\times
\left[
\frac
{q(\fvect | \yvect, \thetavect)}{q(\fvect^{\prime} | \yvect, \thetavect^{\prime})}
\frac
{\pi(\thetavect | \thetavect^{\prime})}{\pi(\thetavect^{\prime} | \thetavect)} 
\right]
\end{equation}
Isolating the terms in the squared bracket allows us to interpret $\tilde{z}$ as a Hastings ratio with a joint proposal for $\thetavect$ and for the importance sample $\fvect$ given by
\begin{equation}
\pi(\fvect^{\prime}, \thetavect^{\prime} | \fvect, \thetavect) = q(\fvect^{\prime} | \yvect, \thetavect^{\prime}) \pi(\thetavect^{\prime} | \thetavect).
\end{equation}
The remaining term in $\tilde{z}$ indicates that the target distribution this approach is sampling from is 
\begin{equation}
p(\yvect | \fvect) p(\fvect | \thetavect) p(\thetavect) = p(\yvect, \fvect, \thetavect).
\end{equation}
If we concentrate on $\thetavect$, regardless of $\fvect$, the target distribution is exactly what we are aiming to sample from, as it is proportional to the posterior $p(\thetavect | \yvect)$.
The extension to more than one importance sample follows from a similar argument, except that the approximating density $q(\fvect | \yvect, \thetavect)$ appears in the expression of the target distribution; however, this does not cause any problems as marginalizing latent variables leads to the same conclusion as in the case $N_{\mathrm{imp}} = 1$.
The analysis for $N_{\mathrm{imp}} = 1$ also reveals an interesting similarity with the approach proposed in~\cite{KnorrHeld02}, where a joint update of $\thetavect$ and $\fvect$ was performed as follows: proposing $\thetavect^{\prime} | \thetavect$, proposing $\fvect^{\prime} | \yvect, \thetavect^{\prime}$, and accepting/rejecting the joint proposal $\thetavect^{\prime},\fvect^{\prime}$. However in this case the PM transition kernel will still target the desired marginal posterior irrespective of the value of importance samples $N_{\mathrm{imp}}$.

\section{Assessing importance distributions}

In this section, we present simulations to assess the ability of the PM approach to characterize the marginal likelihood $p(\yvect | \thetavect)$ in GP classification.
First, we aim to assess the quality of the estimate given by the importance sampler based on LA and EP on simulated data with respect to the number of importance samples.
Second, we will evaluate the efficiency of the sampler on simulated data with respect to the approximation used to draw importance samples and with respect to their number.
Third, we will compare the PM approach with the AA and SURR parameterizations that are the most efficient sampling schemes proposed in the literature for sampling hyper-parameters in models involving GP priors~\cite{FilipponeML13}.
In all the experiments, in both LA and EP we imposed a convergence criterion on the change in squared norm of $\fvect$ being less than $n / 10^4$.

\subsection{Analysis of the variance of the estimator}

\begin{figure}[th]
  \begin{center}
    \begin{tabular}{cc}
         \hspace{-0.5cm} \includegraphics[width=\figwidth,height=\figwidth]{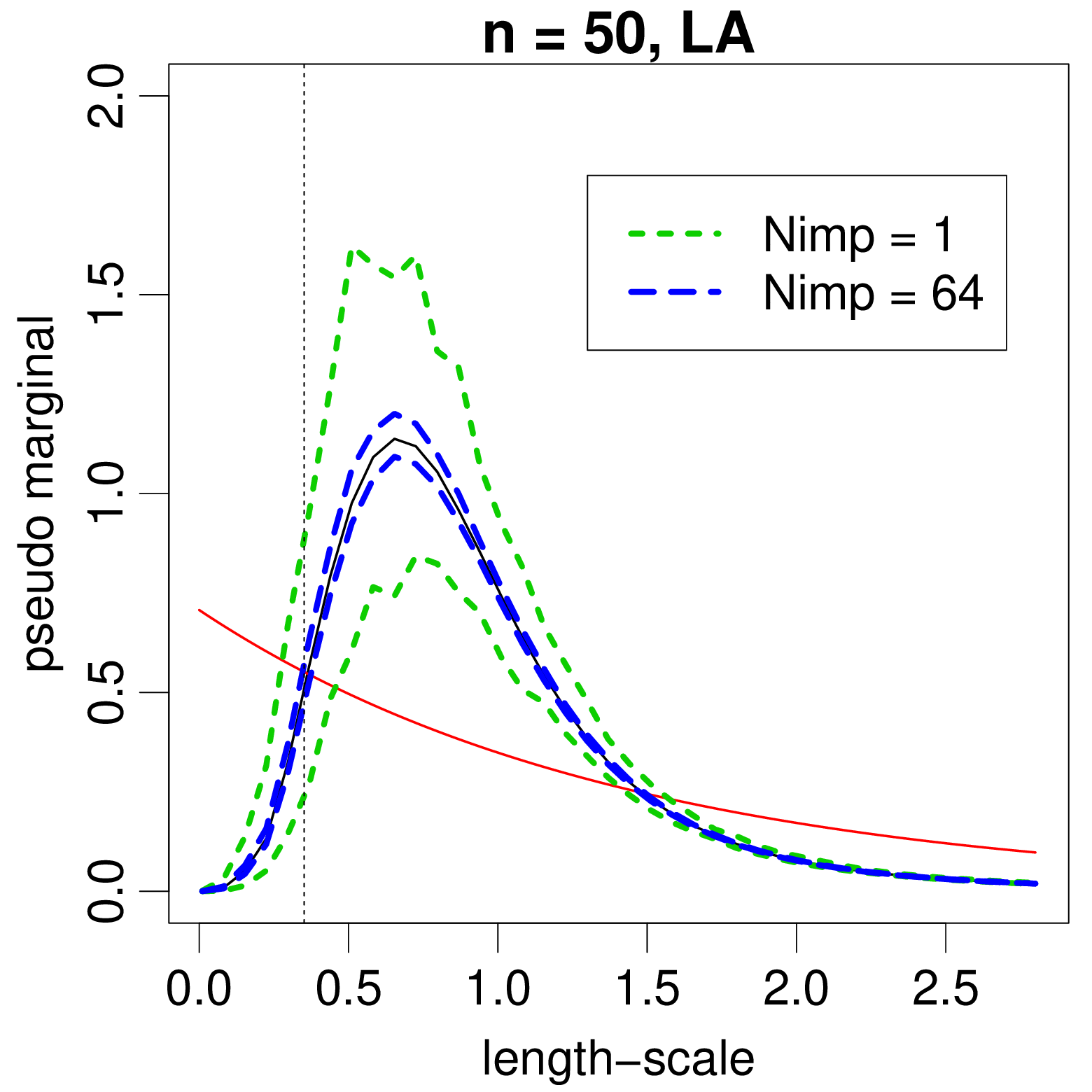}  & \hspace{-0.5cm} \includegraphics[width=\figwidth,height=\figwidth]{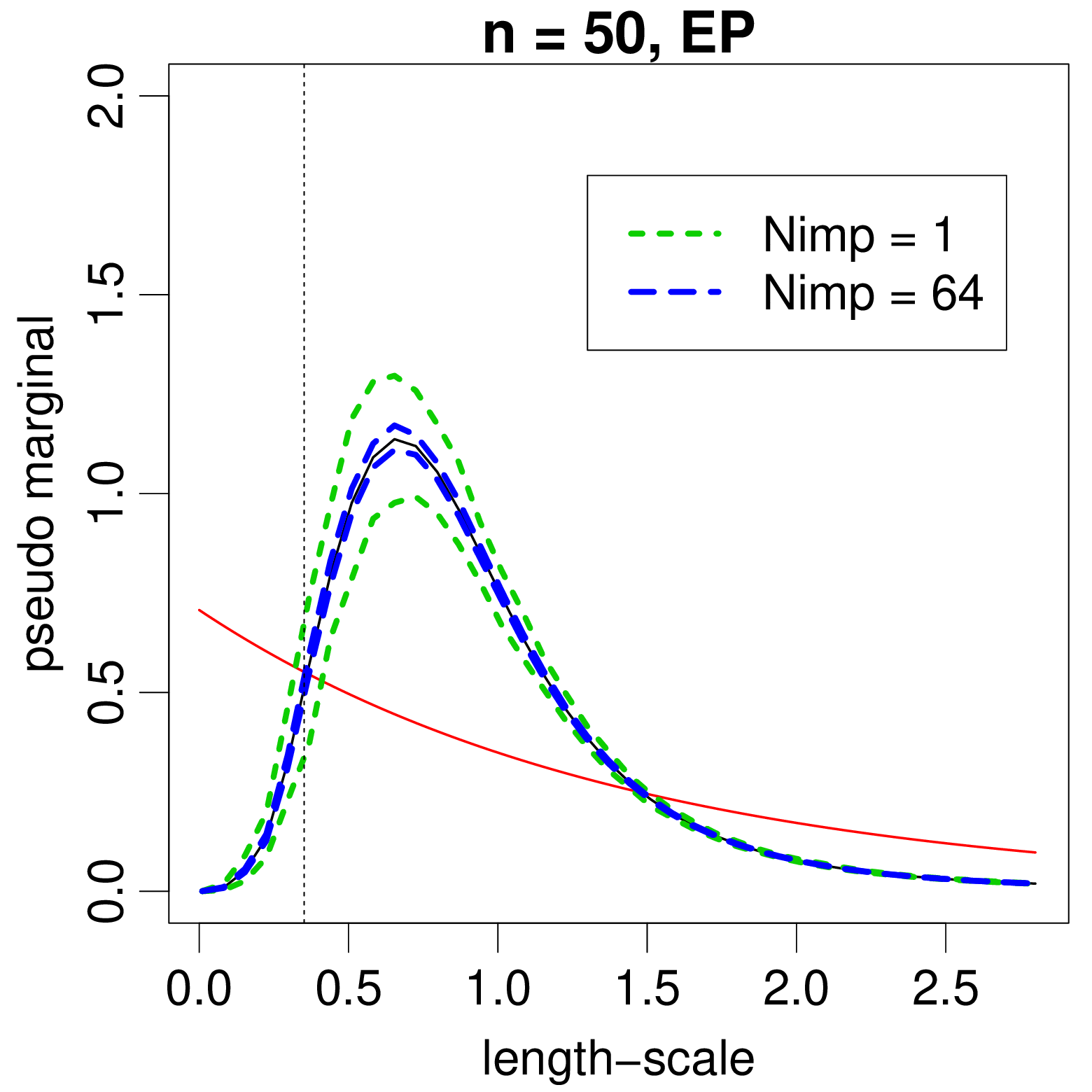} \\
         \hspace{-0.5cm} \includegraphics[width=\figwidth,height=\figwidth]{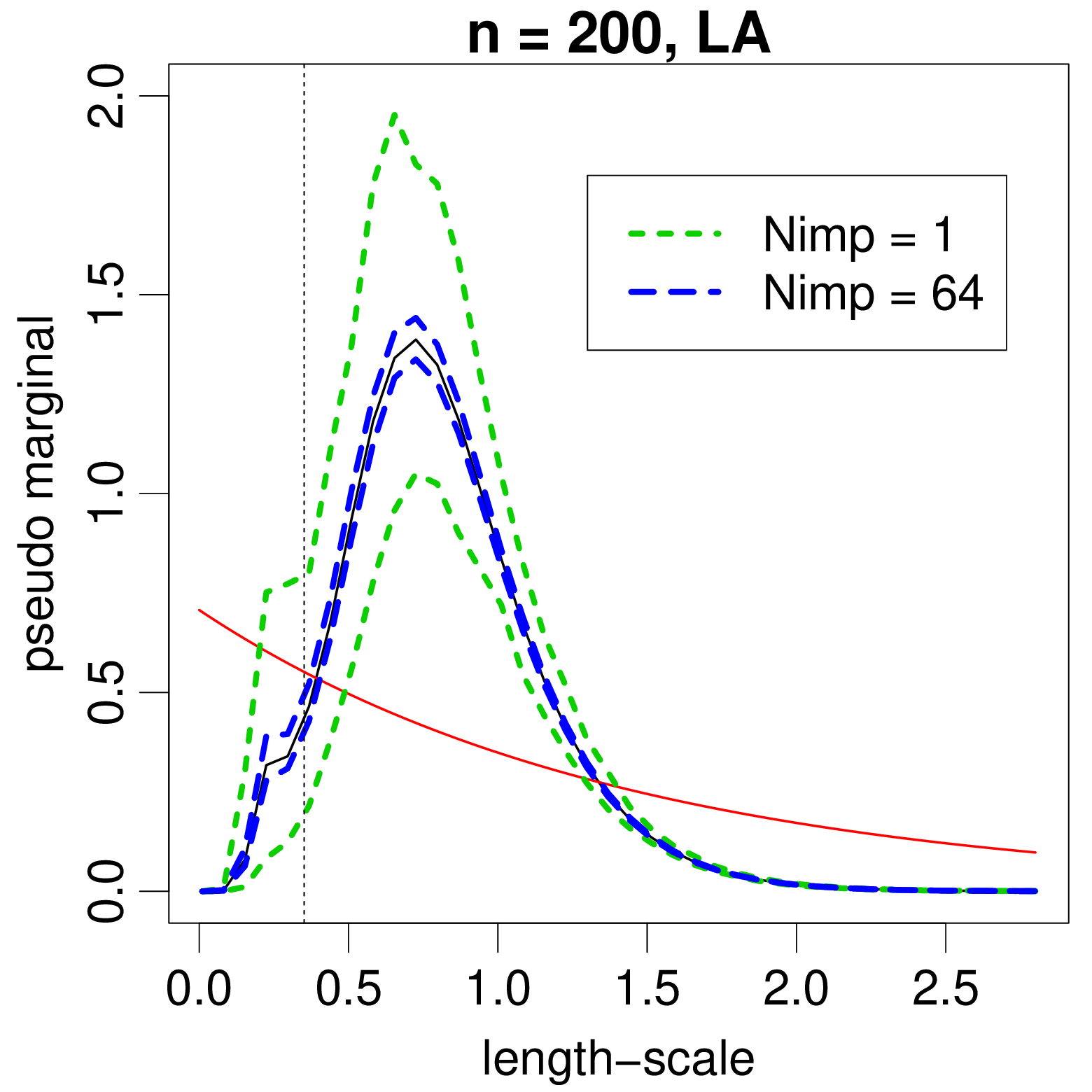}  & \hspace{-0.5cm} \includegraphics[width=\figwidth,height=\figwidth]{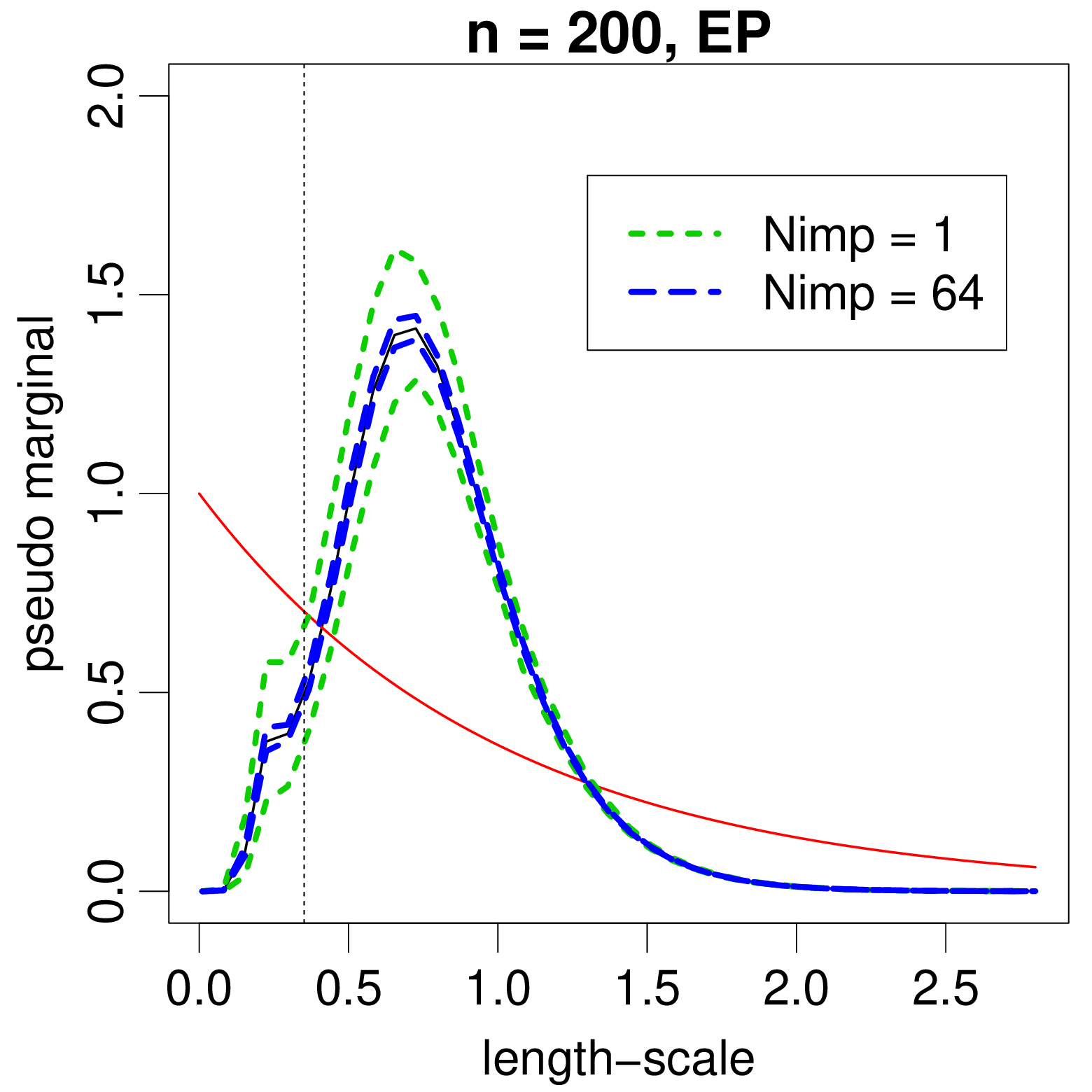} \\
    \end{tabular}
    \caption{Plot of the PM as a function of the length-scale $\tau$; black solid lines represent the average over $500$ repetitions and dashed lines represent 2.5th and 97.5th quantiles for $N_{\mathrm{imp}} = 1$ and $N_{\mathrm{imp}} = 64$. The solid red line is the prior density.}  \label{fig:assess:logreg}
  \end{center}
\end{figure}

In this section, we present an assessment of the variance of the estimator $\tilde{p}(\thetavect | \yvect)$ with respect to the global length-scale parameter $\tau$ in equation~\ref{eq:covariance:isotropic}.
In particular, we are interested in comparing the quality of the approximation given by the importance sampling approach based on the two different approximations employed in this work. \label{REV:AE:5} \label{REV:1:3} \label{REV:2:7} \label{REV:2:17}
Given that the dimensionality of the space where the approximation is performed grows linearly with the number of input vectors $n$, we are also interested in studying the effect of $n$ on the quality of the approximation.
Based on these considerations, we simulated data from the GP classification model with $d=2$ and $n=50$ and $n=200$, with an isotropic covariance function with $\tau=0.35$ and $\sigma=2.08$. 
We fixed the value of $\sigma$ to the value used to generate the data, and we imposed a Gamma prior on the length-scale $p(\tau) = \Ga(\tau | a_{\tau}, b_{\tau})$ with shape $a_{\tau} = 1$ and rate $b_{\tau} = 1/\sqrt{d}$.
We then computed the posterior over $\thetavect$ based on $\tilde{p}(\yvect | \thetavect)$ for different values of $\tau$ and over $500$ repetitions, with different number of importance samples $N_{\mathrm{imp}}$.
The results are reported in figure~\ref{fig:assess:logreg}.


As expected, for larger sets of importance samples, the estimates are more accurate.
We can also see that EP leads to a smaller variance compared to LA, which is not surprising given that the approximation achieved by EP is more accurate~\cite{Kuss05,Nickisch08}.
The experiment suggests that there is little increase in the variance of the estimator for the larger data set.

\subsection{Effect of the pseudo marginal on the efficiency of the sampler}

\begin{table*}[ht]
{
\ifsinglecol
\tiny
\fi
\caption{Analysis of convergence and efficiency of a MH algorithm sampling the hyper-parameters using the PM approach. The results show the dependency of the effective sample size (ESS) and speed of convergence (measured through the $\hat{R}$ statistics after $1e3$, $2e3$, $5e3$, and $1e4$ iterations) with respect to the type of approximation (LA or EP) and the number of importance samples used to compute an unbiased estimate of the marginal likelihood $p(\yvect | \thetavect)$.} \label{res:ess:rhat:approx}
\begin{center}
\begin{tabular}{ccl|cccccc|cccccc}
&&& \multicolumn{6}{c|}{Isotropic} & \multicolumn{6}{c}{ARD} \\
$n$ & $d$ & Scheme                    & $\mathrm{ESS}$ & $\hat{R}       $ & $\hat{R}       $ & $\hat{R}      $ & $\hat{R}$ & Acc       & $\mathrm{ESS}$ & $\hat{R}$ & $\hat{R}$ & $\hat{R}$ & $\hat{R}$ & Acc \\ 
    &     &        $(N_{\mathrm{imp}})$ &               & $        {1e3}$ & $        {2e3}$ &         ${5e3}$ & ${1e4}$ & rate      &                & ${1e3}$ & ${2e3}$ & ${5e3}$ & ${1e4}$ & rate  \\ \hline
\multirow{8}{*}{50} & \multirow{8}{*}{2} & PM LA $(1)$ & $749$ $(73)$ & $1.00$ & $1.00$ & $1.00$ & $1.00$  & $23.9$ $(0.4)$ & $131$ $(41)$ & $1.04$ & $1.05$ & $1.04$ & $1.02$  & $13.0$ $(3.3)$  \\
&& PM LA $(16)$ & $778$ $(61)$ & $1.00$ & $1.00$ & $1.00$ & $1.00$  & $25.1$ $(0.4)$ & $206$ $(21)$ & $1.05$ & $1.04$ & $1.03$ & $1.01$  & $16.7$ $(1.2)$  \\
&& PM LA $(64)$ & $752$ $(76)$ & $1.00$ & $1.00$ & $1.00$ & $1.00$  & $24.8$ $(0.5)$ & $212$ $(38)$ & $1.03$ & $1.03$ & $1.02$ & $1.01$  & $16.6$ $(1.9)$  \\
&& PM EP $(1)$ & $747$ $(100)$ & $1.00$ & $1.00$ & $1.00$ & $1.00$  & $24.4$ $(0.5)$ & $208$ $(22)$ & $1.01$ & $1.00$ & $1.01$ & $1.00$  & $16.1$ $(1.0)$  \\
&& PM EP $(16)$ & $736$ $(163)$ & $1.00$ & $1.00$ & $1.00$ & $1.00$  & $25.0$ $(0.6)$ & $246$ $(23)$ & $1.00$ & $1.00$ & $1.00$ & $1.00$  & $19.2$ $(0.7)$  \\
&& PM EP $(64)$ & $793$ $(54)$ & $1.00$ & $1.00$ & $1.00$ & $1.00$  & $24.9$ $(0.5)$ & $252$ $(24)$ & $1.00$ & $1.00$ & $1.00$ & $1.00$  & $19.9$ $(0.8)$  \\
&& AA & $287$ $(58)$ & $1.00$ & $1.00$ & $1.00$ & $1.00$  & $22.7$ $(0.9)$ & $20$ $(9)$ & $1.12$ & $1.12$ & $1.05$ & $1.03$  & $20.3$ $(2.6)$  \\
&& SURR & $154$ $(11)$ & $1.01$ & $1.00$ & $1.00$ & $1.00$  & $22.4$ $(1.0)$ & $21$ $(6)$ & $1.11$ & $1.08$ & $1.07$ & $1.05$  & $21.6$ $(3.3)$  \\
\hline

\multirow{8}{*}{50} & \multirow{8}{*}{10} & PM LA $(1)$ & $237$ $(147)$ & $1.25$ & $1.26$ & $1.19$ & $1.08$  & $18.2$ $(4.6)$ & $19$ $(12)$ & $1.08$ & $1.09$ & $1.13$ & $1.17$  & $7.1$ $(4.4)$  \\
&& PM LA $(16)$ & $238$ $(194)$ & $1.01$ & $1.02$ & $1.03$ & $1.02$  & $16.9$ $(6.0)$ & $62$ $(30)$ & $1.04$ & $1.04$ & $1.06$ & $1.11$  & $17.6$ $(4.9)$  \\
&& PM LA $(64)$ & $348$ $(126)$ & $1.01$ & $1.01$ & $1.01$ & $1.01$  & $21.8$ $(2.3)$ & $78$ $(23)$ & $1.03$ & $1.03$ & $1.01$ & $1.01$  & $20.2$ $(2.8)$  \\
&& PM EP $(1)$ & $282$ $(47)$ & $1.02$ & $1.01$ & $1.01$ & $1.00$  & $19.2$ $(1.4)$ & $63$ $(18)$ & $1.03$ & $1.03$ & $1.02$ & $1.01$  & $15.6$ $(1.2)$  \\
&& PM EP $(16)$ & $507$ $(36)$ & $1.00$ & $1.00$ & $1.00$ & $1.00$  & $24.7$ $(1.0)$ & $107$ $(12)$ & $1.01$ & $1.01$ & $1.01$ & $1.01$  & $24.2$ $(0.5)$  \\
&& PM EP $(64)$ & $583$ $(51)$ & $1.00$ & $1.00$ & $1.00$ & $1.00$  & $26.1$ $(0.6)$ & $108$ $(28)$ & $1.02$ & $1.01$ & $1.01$ & $1.01$  & $26.3$ $(0.7)$  \\
&& AA & $71$ $(23)$ & $1.02$ & $1.01$ & $1.02$ & $1.01$  & $22.8$ $(2.5)$ & $74$ $(6)$ & $1.03$ & $1.03$ & $1.02$ & $1.02$  & $22.4$ $(2.7)$  \\
&& SURR & $52$ $(15)$ & $1.04$ & $1.02$ & $1.02$ & $1.02$  & $21.8$ $(2.9)$ & $45$ $(5)$ & $1.03$ & $1.03$ & $1.02$ & $1.01$  & $21.7$ $(1.7)$  \\
\hline

\multirow{8}{*}{200} & \multirow{8}{*}{2} & PM LA $(1)$ & $717$ $(31)$ & $1.00$ & $1.00$ & $1.00$ & $1.00$  & $31.7$ $(0.5)$ & $318$ $(29)$ & $1.00$ & $1.00$ & $1.00$ & $1.00$  & $38.3$ $(0.8)$  \\
&& PM LA $(16)$ & $739$ $(46)$ & $1.00$ & $1.00$ & $1.00$ & $1.00$  & $32.6$ $(0.5)$ & $364$ $(20)$ & $1.00$ & $1.00$ & $1.00$ & $1.00$  & $43.1$ $(0.5)$  \\
&& PM LA $(64)$ & $730$ $(26)$ & $1.00$ & $1.00$ & $1.00$ & $1.00$  & $32.6$ $(0.6)$ & $355$ $(42)$ & $1.00$ & $1.00$ & $1.00$ & $1.00$  & $43.6$ $(0.7)$  \\
&& PM EP $(1)$ & $736$ $(47)$ & $1.00$ & $1.00$ & $1.00$ & $1.00$  & $32.8$ $(0.5)$ & $349$ $(26)$ & $1.00$ & $1.00$ & $1.00$ & $1.00$  & $42.0$ $(0.4)$  \\
&& PM EP $(16)$ & $736$ $(48)$ & $1.00$ & $1.00$ & $1.00$ & $1.00$  & $32.7$ $(0.4)$ & $365$ $(21)$ & $1.00$ & $1.00$ & $1.00$ & $1.00$  & $43.5$ $(0.5)$  \\
&& PM EP $(64)$ & $721$ $(43)$ & $1.00$ & $1.00$ & $1.00$ & $1.00$  & $32.6$ $(0.7)$ & $354$ $(37)$ & $1.00$ & $1.00$ & $1.00$ & $1.00$  & $43.9$ $(0.6)$  \\
&& AA & $112$ $(49)$ & $1.01$ & $1.01$ & $1.01$ & $1.01$  & $21.6$ $(1.0)$ & $41$ $(8)$ & $1.07$ & $1.07$ & $1.06$ & $1.04$  & $23.0$ $(2.3)$  \\
&& SURR & $54$ $(8)$ & $1.01$ & $1.02$ & $1.01$ & $1.01$  & $21.5$ $(1.5)$ & $61$ $(9)$ & $1.02$ & $1.02$ & $1.01$ & $1.01$  & $22.0$ $(1.9)$  \\
\hline

\multirow{8}{*}{200} & \multirow{8}{*}{10} & PM LA $(1)$ & $115$ $(53)$ & $1.03$ & $1.03$ & $1.02$ & $1.04$  & $27.1$ $(11.3)$ & $27$ $(10)$ & $1.39$ & $1.33$ & $1.19$ & $1.11$  & $12.3$ $(2.9)$  \\
&& PM LA $(16)$ & $117$ $(42)$ & $1.05$ & $1.04$ & $1.03$ & $1.03$  & $26.8$ $(9.0)$ & $53$ $(18)$ & $1.08$ & $1.06$ & $1.03$ & $1.02$  & $18.0$ $(1.9)$  \\
&& PM LA $(64)$ & $145$ $(62)$ & $1.01$ & $1.01$ & $1.01$ & $1.02$  & $28.7$ $(8.0)$ & $37$ $(31)$ & $1.14$ & $1.15$ & $1.17$ & $1.17$  & $13.2$ $(8.6)$  \\
&& PM EP $(1)$ & $75$ $(35)$ & $1.03$ & $1.03$ & $1.03$ & $1.02$  & $18.2$ $(3.7)$ & $26$ $(14)$ & $1.12$ & $1.09$ & $1.13$ & $1.08$  & $10.3$ $(3.1)$  \\
&& PM EP $(16)$ & $137$ $(38)$ & $1.01$ & $1.01$ & $1.01$ & $1.01$  & $27.6$ $(4.2)$ & $52$ $(13)$ & $1.20$ & $1.15$ & $1.08$ & $1.04$  & $17.1$ $(2.1)$  \\
&& PM EP $(64)$ & $130$ $(64)$ & $1.09$ & $1.09$ & $1.09$ & $1.12$  & $25.5$ $(10.0)$ & $45$ $(21)$ & $1.03$ & $1.03$ & $1.03$ & $1.09$  & $17.4$ $(3.4)$  \\
&& AA & $26$ $(6)$ & $1.05$ & $1.04$ & $1.03$ & $1.03$  & $23.7$ $(1.9)$ & $22$ $(7)$ & $1.10$ & $1.08$ & $1.05$ & $1.03$  & $21.6$ $(5.5)$  \\
&& SURR & $18$ $(7)$ & $1.27$ & $1.28$ & $1.24$ & $1.16$  & $21.5$ $(8.6)$ & $10$ $(3)$ & $1.07$ & $1.08$ & $1.06$ & $1.05$  & $20.3$ $(3.9)$  \\
\hline

\end{tabular}
\end{center}
}
\end{table*}

In this section we report an analysis on simulated data showing how the choice of the approximation and the number of importance samples affect the efficiency in sampling from $p(\thetavect | \yvect)$.
We generated data sets from the GP classification model with different combinations of number of input vectors and number of covariates.
The covariates were generated in the unit hypercube and data were selected to have an equal number of input vectors in each class.
We chose Gamma priors for the hyper-parameters as follows: $p(\tau_i) = \Ga(\tau_i | a_{\tau}, b_{\tau})$ with shape $a_{\tau} = 1$ and rate $b_{\tau} = 1/\sqrt{d}$, and $p(\sigma) = \Ga(\sigma | a_{\sigma}, b_{\sigma})$ with shape $a_{\sigma} = 1.2$ and rate $b_{\sigma} = 0.2$.
In the formulation of the GP classification model, all hyper-parameters have to be positive; for the sake of convenience, we reparameterized them introducing the variables $\psi_{\tau_i} = \log(\tau_i)$ and $\psi_{\sigma} = \log(\sigma)$.

For each method, we ran $10$ parallel chains for $5000$ burn-in iterations followed by $10000$ iterations; convergence speed of the samplers was monitored using the Potential Scale Reduction Factor (PSRF) ($\hat{R}$ statistics) as described in~\cite{Gelman92}.
The chains were initialized from the prior, rather than using the procedure suggested in~\cite{Gelman92} to make the convergence test more challenging.
Also, correctness of the code was checked by using the idea presented in~\cite{Geweke04}, that indirectly shows that the Markov chains have indeed $p(\thetavect | \yvect)$ as their invariant distribution.


The proposal mechanism $\pi(\thetavect^{\prime} | \thetavect)$ was the same for all the PM approaches for a given combination of $n$ and $d$, so that it is meaningful to analyze the effect of $N_{\mathrm{imp}}$ on sampling efficiency, convergence speed, and acceptance rate. \label{REV:2:18}
In particular, a large variance for the estimator of the marginal likelihood can eventually lead to the acceptance of $\thetavect$ because $p(\yvect | \thetavect)$ is largely overestimated leading to a difficulty for the chain to move away from there.
In this case, the chain can get stuck and take several iterations before moving again; this effect has been reported in~\cite{Andrieu09,Beaumont03}.
To isolate the effect of $N_{\mathrm{imp}}$ and the type of approximation on sampling efficiency and acceptance rate, we tuned the chains using preliminary runs for EP and $N_{\mathrm{imp}} = 64$ to achieve about $25\%$ acceptance rate and used the same proposal for LA and other values of $N_{\mathrm{imp}}$.

The results are reported in table \ref{res:ess:rhat:approx} for isotropic and ARD covariances.
As a measure of efficiency, we used the minimum Effective Sample Size (ESS)~\cite{Gilks96} across the hyper-parameters.
The tables also report the median of $\hat{R}$ achieved by the chains at different iterations, namely $1000$, $2000$, $5000$, and $10000$.
This gives an idea of the convergence as the iterations progress. 
Finally, in table \ref{res:ess:rhat:approx} we report the acceptance rate; a low acceptance rate compared to the one obtained by PM~EP~$(64)$ indicates that the chains are more likely to get stuck due to a large variance of the estimator of the marginal likelihood.


The results indicate that sampling efficiency when employing EP to approximate the posterior distribution over $\fvect$ is higher than when employing the LA algorithm.
It is striking to see that evaluating the PM with as little as one importance sample seems already able to offer acceptable performance in terms of ESS compared to larger values of $N_{\mathrm{imp}}$.
However, a low acceptance rate when $N_{\mathrm{imp}}$ is small suggests that the corresponding chains can take several iterations before accepting any proposals.

\begin{table}[ht]
{
\ifsinglecol
\tiny
\fi
\caption{Average number of operations in $O(n^3)$ required for each iteration of the PM approaches with the LA and EP approximations and for each iteration in the AA and SURR parameterizations.} \label{res:complexity:pm:aa:surr}
\begin{center}
\begin{tabular}{ccl|c|c}
$n$ & $d$ & Scheme                    & Isotropic     & ARD \\
\hline

\multirow{4}{*}{50} & \multirow{4}{*}{2} 
 & PM LA & $8.2$ $(0.2)$  & $8.8$ $(0.2)$  \\
&& PM EP & $16.3$ $(0.0)$ & $17.7$ $(0.1)$  \\
&& AA    & $1.0$ $(0.0)$  & $1.0$ $(0.0)$  \\
&& SURR  & $3.0$ $(0.0)$  & $3.0$ $(0.0)$  \\

\hline

\multirow{4}{*}{50} & \multirow{4}{*}{10} 
 & PM LA  & $7.8$ $(0.3)$ & $8.3$ $(0.2)$  \\
&& PM EP  & $13.0$ $(0.1)$ & $13.7$ $(0.1)$  \\
&& AA     & $1.0$ $(0.0)$ & $1.0$ $(0.0)$  \\
&& SURR   & $3.0$ $(0.0)$ & $3.0$ $(0.0)$  \\

\hline

\multirow{4}{*}{200} & \multirow{4}{*}{2} 
 & PM LA  & $8.7$ $(0.0)$ & $9.3$ $(0.0)$  \\
&& PM EP  & $15.5$ $(0.0)$ & $17.3$ $(0.0)$  \\
&& AA     & $1.0$ $(0.0)$ & $1.0$ $(0.0)$  \\
&& SURR   & $3.0$ $(0.0)$ & $3.0$ $(0.0)$  \\

\hline

\multirow{4}{*}{200} & \multirow{4}{*}{10} 
 & PM LA  & $7.6$ $(0.2)$ & $7.9$ $(0.2)$ \\
&& PM EP  & $13.3$ $(0.1)$ & $13.2$ $(0.2)$ \\
&& AA     & $1.0$ $(0.0)$ & $1.0$ $(0.0)$ \\
&& SURR   & $3.0$ $(0.0)$ & $3.0$ $(0.0)$ \\

\hline

\end{tabular}
\end{center}
}
\end{table}

\subsection{Comparison with reparameterization techniques}


Table~\ref{res:ess:rhat:approx} also reports a comparison of the PM method with the AA and SURR sampling schemes with a Metropolis-Hastings transition operator so that results are meaningfully comparable. \label{REV:2:8}
The proposal mechanism was tuned during the burn-in phase to achieve about $25\%$ acceptance rate.
Table~\ref{res:ess:rhat:approx} shows that the PM approach achieves faster convergence and higher sampling efficiency compared to the AA scheme.
The SURR method has comparable convergence behavior and efficiency compared to the AA scheme.
This is somehow different from what presented in~\cite{Murray10}, where a component-wise slice sampling transition operator was employed.
In~\cite{Murray10}, the SURR method achieved higher efficiency per covariance construction, likelihood evaluation and running time compared to the AA method.
In the experiment reported here ESS is comparable.

Table~\ref{res:complexity:pm:aa:surr} reports the average number of operations in $O(n^3)$ needed for one iteration of the PM approach with LA and EP approximations and the AA and SURR parameterizations. \label{REV:1:4} \label{REV:1:12} \label{REV:2:11}
The table shows that EP is more expensive than the LA algorithm, and that the PM approaches require more $O(n^3)$ operations compared to AA and SURR.
Normalization of the ESS by the number of operations suggests that the cost of obtaining independent samples using the PM approach is generally better than AA and SURR.
In the PM approach we also tried stopping the approximation algorithms using a looser convergence criterion (results not reported); especially for $N_{\mathrm{imp}} = 64$ this yielded similar efficiency with much lower computational cost.

\begin{table*}[ht]
{
\ifsinglecol
\tiny
\fi
\caption{Analysis of convergence and efficiency of the AA and SURR parameterizations compared to the PM approach. Each of the Gibbs sampling updates was repeated $20$ times in order to highlight the effect of the parameterization alone. The PM approach is based on the EP approximation and $N_{\mathrm{imp}} = 64$. 
The column next to the one reporting the $\mathrm{ESS}$ shows the $\mathrm{ESS}$ normalized by the number of operations in $O(n^3)$.
} \label{res:ess:rhat:approx:push}
\begin{center}
\begin{tabular}{ccl|cccccc|cccccc}
&&& \multicolumn{6}{c|}{Isotropic} & \multicolumn{6}{c}{ARD} \\
$n$ & $d$ & Scheme                    & $\mathrm{ESS}$ & $\mathrm{ESS}$ & $\hat{R}       $ & $\hat{R}       $ & $\hat{R}      $ & $\hat{R}$ & $\mathrm{ESS}$ & $\mathrm{ESS}$ & $\hat{R}$ & $\hat{R}$ & $\hat{R}$ & $\hat{R}$ \\ 
    &     &                           &                & $/O(n^3)$      & $        {1e3}$ & $        {2e3}$ &         ${5e3}$ & ${1e4}$ &         &   $/O(n^3)$     & ${1e3}$ & ${2e3}$ & ${5e3}$ & ${1e4}$ \\ \hline
\multirow{3}{*}{50} & \multirow{3}{*}{2} & PM   &   $9057$ $(243)$ & $556$ $(15)$   & $1.00$ & $1.00$ & $1.00$ & $1.00$  & $4301$ $(169)$ & $243$ $(10)$ & $1.00$ & $1.00$ & $1.00$ & $1.00$  \\
                    &                    & AA   &   $1571$ $(378)$ & $1571$ $(378)$ & $1.00$ & $1.00$ & $1.00$ & $1.00$  & $57$ $(11)$    &  $57$ $(11)$ & $1.01$ & $1.01$ & $1.01$ & $1.01$  \\
                    &                    & SURR &   $702$ $(98)$   & $234$ $(33)$   & $1.00$ & $1.00$ & $1.00$ & $1.00$  & $64$ $(9)$     &  $21$ $(3)$  & $1.02$ & $1.02$ & $1.02$ & $1.01$  \\
\hline                                                                                                                                                  
                                                                                                                                                        
\multirow{3}{*}{50} & \multirow{3}{*}{10} & PM   &  $7478$ $(333)$ & $575$ $(26)$   & $1.00$ & $1.00$ & $1.00$ & $1.00$ & $2207$ $(141)$  & $161$ $(10)$ & $1.00$ & $1.00$ & $1.00$ & $1.00$  \\
                    &                     & AA   &  $211$ $(22)$   & $211$ $(22)$   & $1.01$ & $1.00$ & $1.00$ & $1.00$ & $464$ $(40)$    & $464$ $(40)$ & $1.00$ & $1.00$ & $1.00$ & $1.00$  \\
                    &                     & SURR &  $133$ $(23)$   & $44$ $(8)$     & $1.01$ & $1.01$ & $1.01$ & $1.00$ & $207$ $(18)$    &  $69$  $(6)$ & $1.01$ & $1.01$ & $1.01$ & $1.00$  \\
\hline                                                                                                                                                  
                                                                                                                                                        
\multirow{3}{*}{200} & \multirow{3}{*}{2} & PM   &  $8850$ $(301)$ & $571$ $(19)$   & $1.00$ & $1.00$ & $1.00$ & $1.00$ & $5844$ $(166)$  & $338$ $(10)$ & $1.00$ & $1.00$ & $1.00$ & $1.00$  \\
                     &                    & AA   &  $327$ $(142)$  & $327$ $(142)$  & $1.01$ & $1.00$ & $1.01$ & $1.00$ & $143$ $(22)$    & $143$ $(22)$ & $1.01$ & $1.01$ & $1.00$ & $1.00$  \\
                     &                    & SURR &  $234$ $(24)$   & $78$ $(8)$     & $1.00$ & $1.00$ & $1.00$ & $1.00$ & $339$ $(47)$    & $113$ $(16)$ & $1.00$ & $1.00$ & $1.00$ & $1.00$  \\
\hline                                                                                                                                                  
                                                                                                                                                        
\multirow{3}{*}{200} & \multirow{3}{*}{10} & PM   & $737$ $(533)$  & $55 (40)$      & $1.01$ & $1.01$ & $1.01$ & $1.01$ & $363$ $(250)$   &  $28$ $(19)$ & $1.03$ & $1.03$ & $1.01$ & $1.01$  \\
                     &                     & AA   & $37$ $(7)$     & $37$ $(7)$     & $1.05$ & $1.05$ & $1.05$ & $1.03$ & $42$ $(8)$      &  $42$  $(8)$ & $1.03$ & $1.03$ & $1.02$ & $1.01$  \\
                     &                     & SURR & $25$ $(6)$     & $8$ $(2)$      & $1.05$ & $1.05$ & $1.01$ & $1.00$ & $22$ $(5)$      &   $7$  $(2)$ & $1.09$ & $1.07$ & $1.05$ & $1.03$  \\
\hline

\end{tabular}
\end{center}
}
\end{table*}

We conducted a further test of convergence with the aim of mitigating the effect of the random walk exploration and highlighting the advantages offered by different parameterizations.
We ran the AA, SURR and PM approaches by repeating each step of the Gibbs sampler $20$ times, so that we can roughly consider the new sample drawn in a Gibbs sampling updates independent with respect to the previous.
The behavior of the $\hat{R}$ statistics with respect to the number of iterations, reported in table \ref{res:ess:rhat:approx:push}, reveals some interesting features.
All methods are characterized by fast convergence.
In the case of the SURR and AA methods, however, efficiency is much lower than what can be achieved by the PM method when repeating the Gibbs sampling update $\thetavect | \yvect$ $20$ times. 
This is an indication that the parameterizations of SURR and AA methods are not fully capable of breaking the correlation between hyper-parameters and latent variables.
Finally, note that in the case of ARD covariances, the low efficiency in all methods is due to the random walk type of exploration; in those cases $20$ iterations of the Gibbs sampling steps were not enough to ensure independence between consecutive samples.

\subsection{Convergence speed and efficiency on real data}

This section reports a comparison of classification performance on three UCI data sets~\cite{Asuncion07}, so as to verify the capability of the proposed approach to effectively carry out inference for parameters in general GP classification problems.
The Breast and Pima data sets are two-class classification problems, described by $d=9$ and $d=8$ covariates, comprising $n=682$ and $n=768$ input vectors respectively.
The Abalone data set has three classes; in this paper, we considered the task of inferring parameters of a GP probit classifier for the two classes ``M'' and ``F'', resulting in a data set of $n=2835$ input vectors and $d=8$ covariates.
All covariates were transformed to have zero mean and unit standard deviation.
We selected the isotropic covariance function in equation~\ref{eq:covariance:isotropic} and chose the following priors for their parameters: $p(\tau) = \Ga(\tau | a_{\tau}, b_{\tau})$ with shape $a_{\tau} = 1$ and rate $b_{\tau} = 1/\sqrt{d}$, and $p(\sigma) = \Ga(\sigma | a_{\sigma}, b_{\sigma})$ with shape $a_{\sigma} = 1.1$ and rate $b_{\sigma} = 0.1$.

We compared the proposed sampling PM approach with the AA and SURR parameterizations.
In the latter two, we alternated the sampling of $\thetavect$ using the MH algorithm and the sampling of $\fvect$ iterating ELL-SS ten times.
In the PM approach we selected an approximation based on the LA algorithm and chose the number of importance samples to be $N_{\mathrm{imp}} = 1$.

We ran chains for $12000$ iterations, where the first $2000$ were used to tune the proposal mechanisms to achieve an acceptance rate between $20\%$ and $30\%$.
In the case of the PM approach, in the adaptive phase we used the approximate marginal likelihood obtained by the LA algorithm.
This was to overcome the problems that may arise when chains get trapped due to a largely overestimated value of the marginal likelihood. 
After $2000$ iterations, we then switched to the estimate of the marginal likelihood obtained by importance sampling.
For each sampling approach, we ran $10$ parallel chains initialized from the prior, so that we could compare convergence speed.

The results for the three UCI data sets are reported in figures~\ref{fig:sampling:breast:pima} and~\ref{fig:sampling:abalone}.
The plots show efficiency and convergence speed in the sampling of the logarithm of the length-scale parameter $\tau$.
The left and middle panels of these figures show the trace and the auto-correlation plots of one chain after the burn-in period of $2000$ iterations.
The auto-correlation plot gives an indication of efficiency; the faster the auto-correlation of a chain reaches values close to zero, the higher the efficiency of the corresponding MCMC approach.
In order to facilitate visualization, the traces were thinned by a factor of $10$ and the auto-correlation plots were computed on the thinned chains.
The right panel shows the evolution of the PSRF ($\hat{R}$) after the burn-in period without thinning the chains.

\ifsinglecol
\begin{figure}[pht]
\else
\begin{figure}[ht]
\fi
  \begin{center}
  {\bf Breast $n = 682$ \vspace{0.2cm}}
     \begin{tabular}{ccc}
      \hspace{-0.4cm} \includegraphics[width=\halffigwidth]{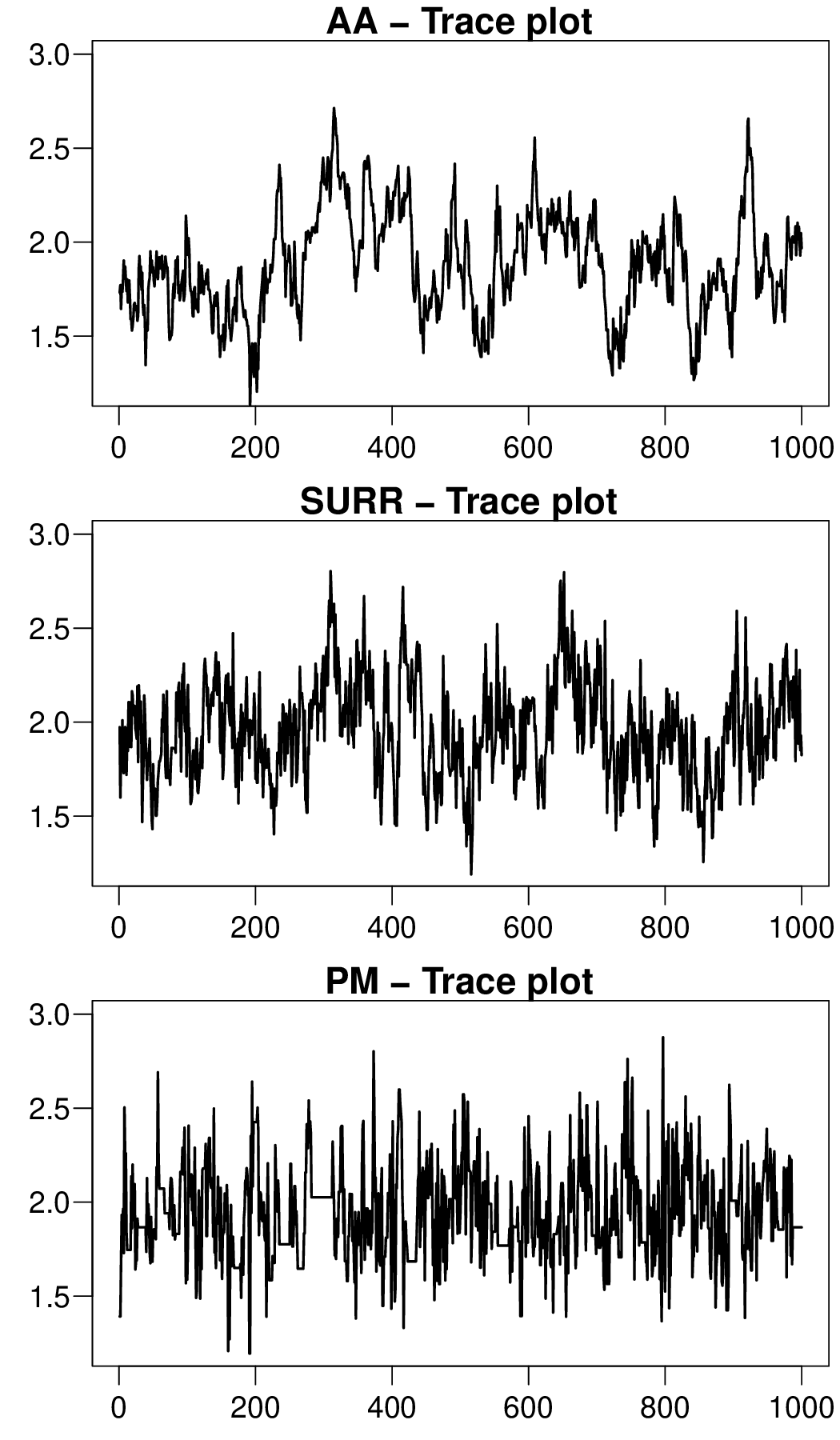}  & \hspace{-0.5cm} \includegraphics[width=\quarterfigwidth]{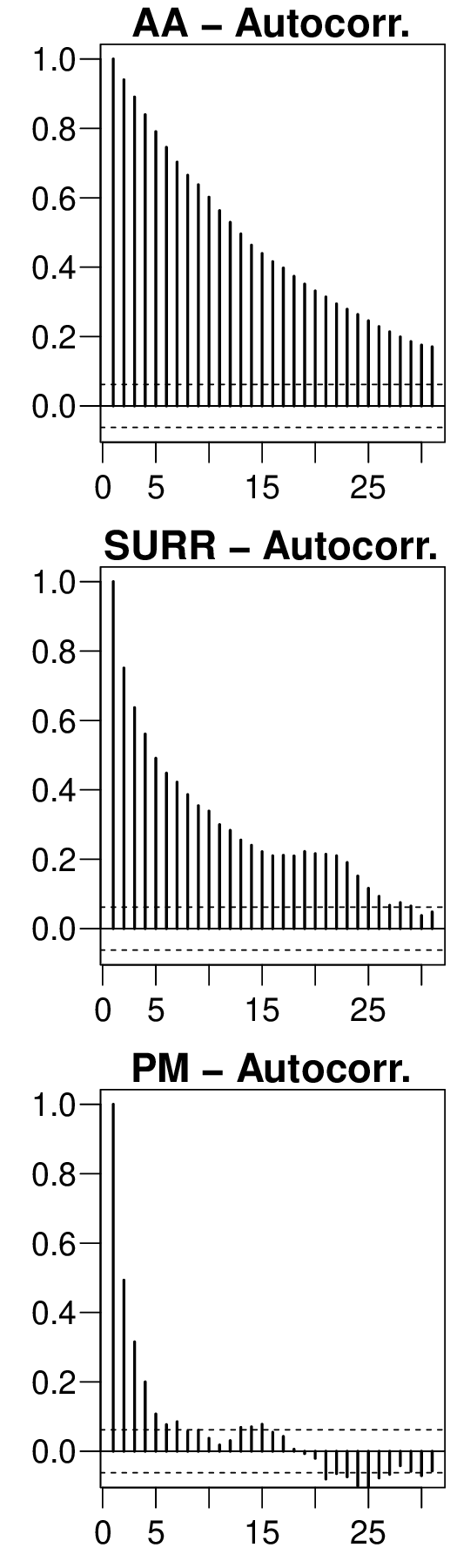} & \hspace{-0.5cm} \includegraphics[width=\quarterfigwidth]{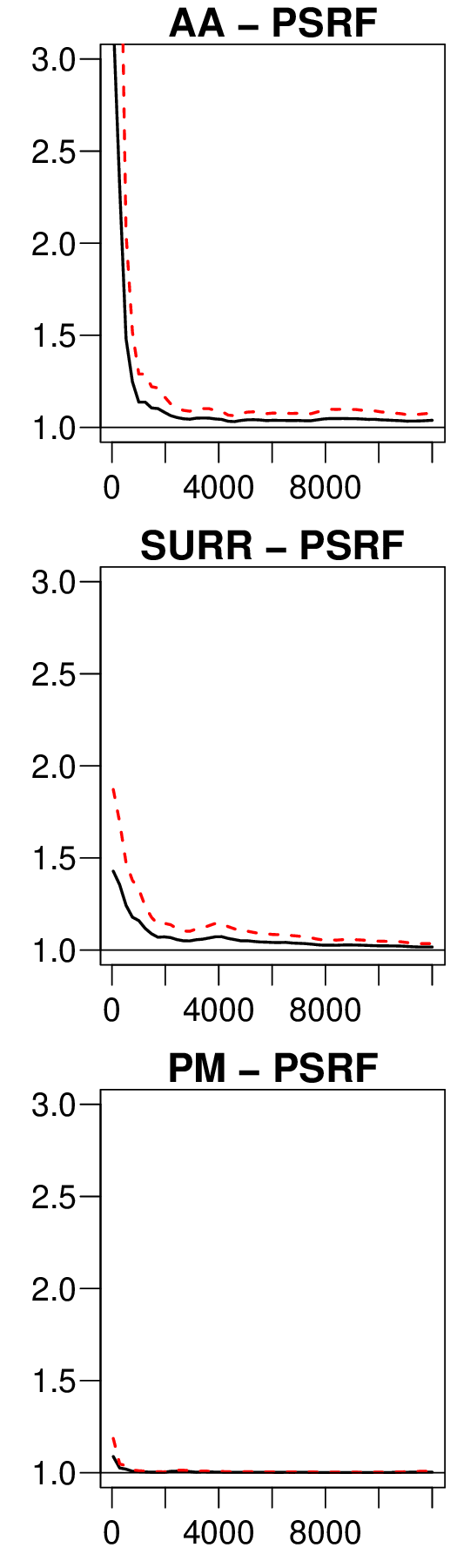} \\
    \end{tabular}

    \vspace{0.3cm}
  {\bf Pima $n = 768$ \vspace{0.2cm}}
     \begin{tabular}{ccc}
      \hspace{-0.4cm} \includegraphics[width=\halffigwidth]{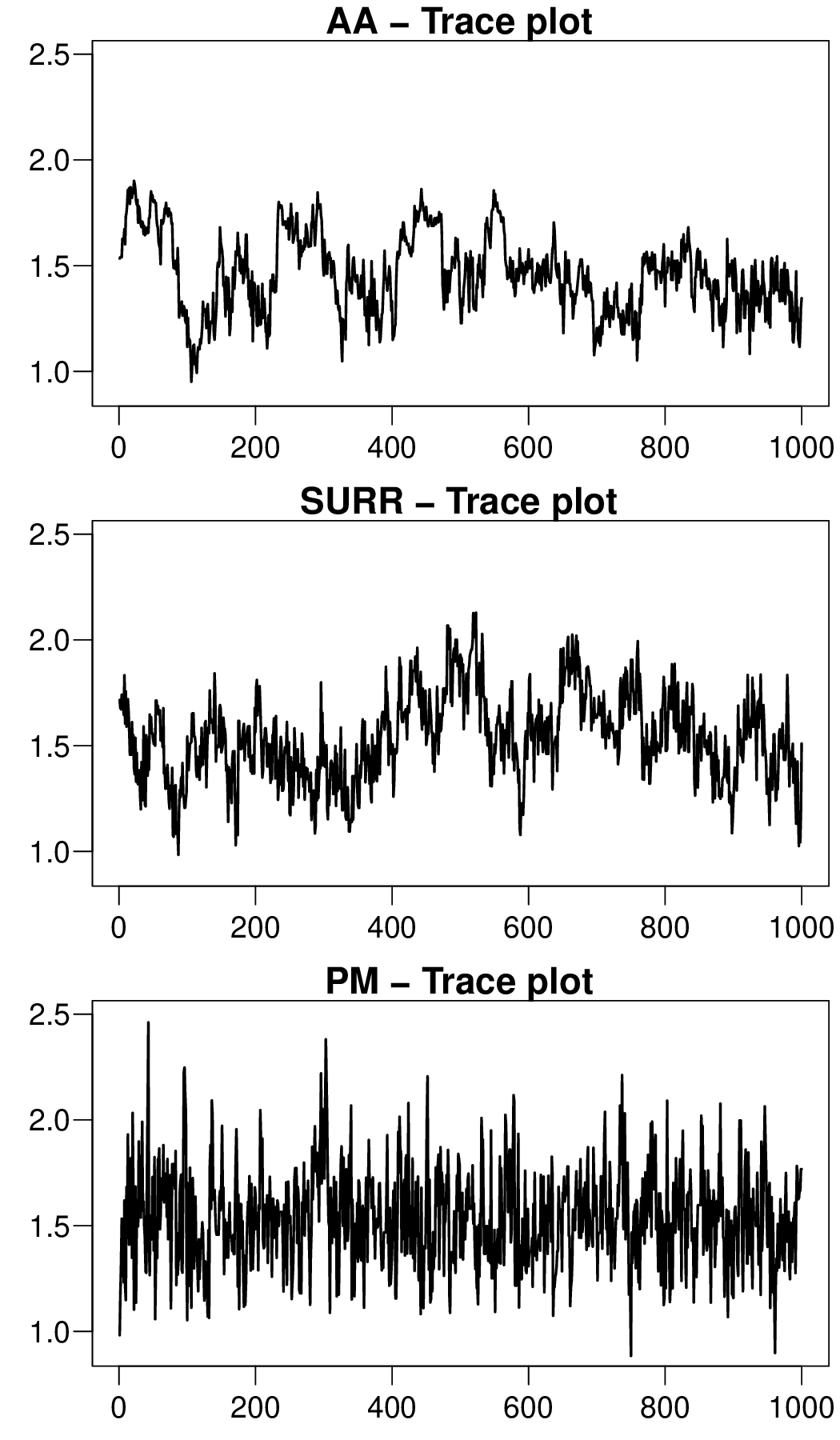}  & \hspace{-0.5cm} \includegraphics[width=\quarterfigwidth]{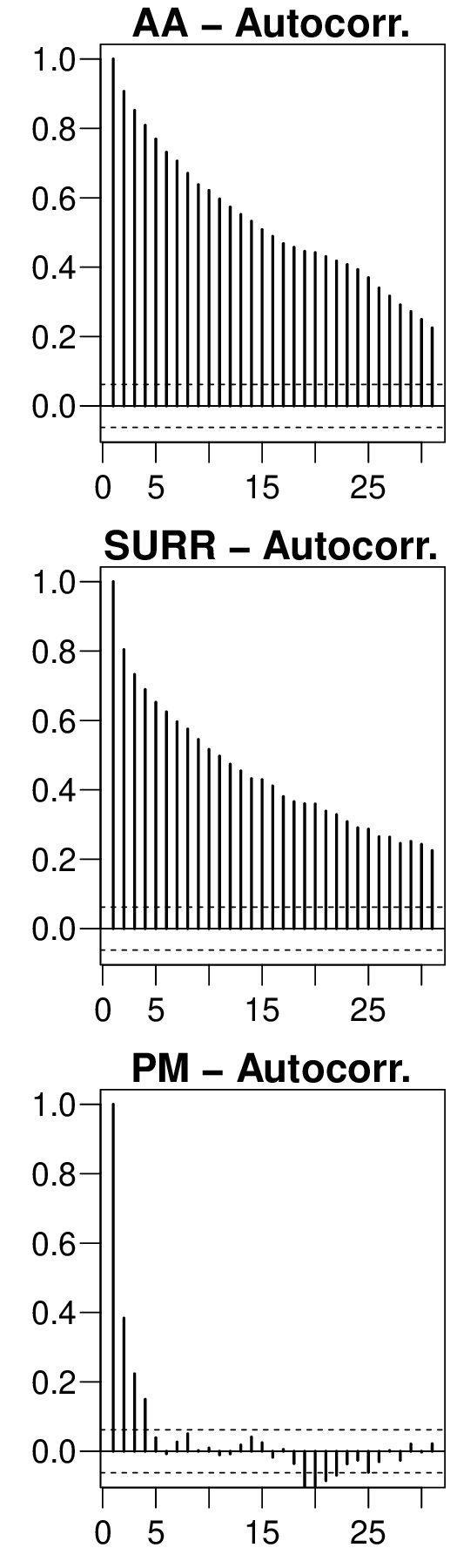} & \hspace{-0.5cm} \includegraphics[width=\quarterfigwidth]{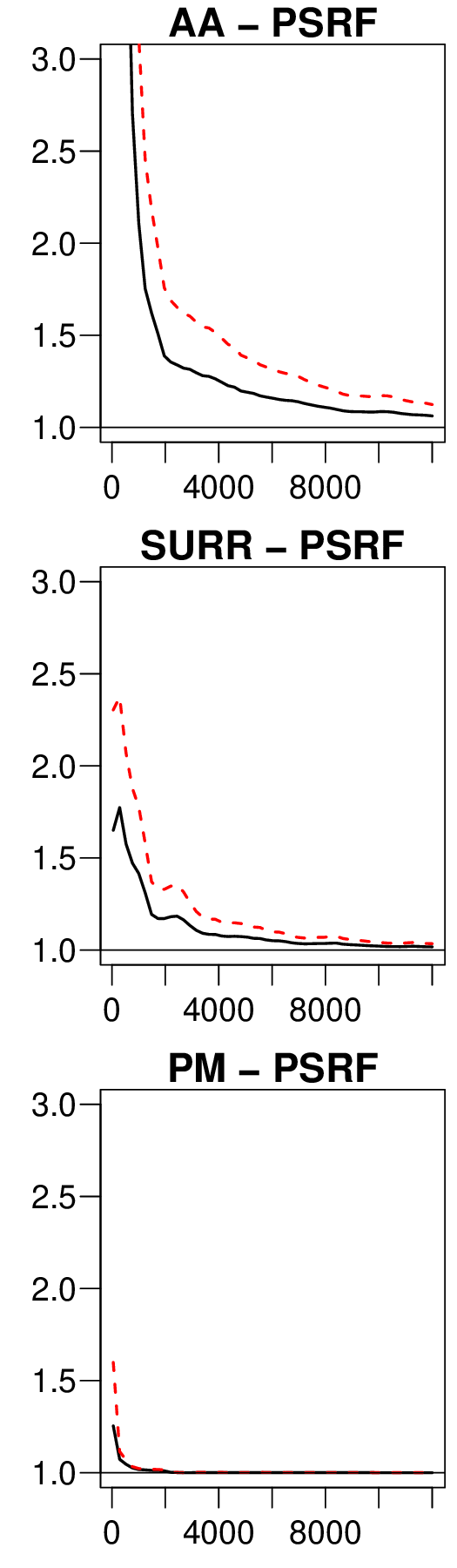} \\
    \end{tabular}
  \end{center}
  \caption{
Summary of efficiency and convergence speed on Breast and Pima data sets for the AA and SURR parameterization and the proposed PM approach.
All plots show the sampling of the logarithm of the length-scale parameter $\tau$.
The figure shows trace plots (left panels) and corresponding auto-correlation plots (middle panels) for one chain thinned by a factor of $10$ after burn-in.
The right panel reports the evolution of the PSRF after burn-in; in this plot the solid line and the red dashed line represent the median and the $97.5\%$ percentile respectively.
}  \label{fig:sampling:breast:pima}
\end{figure}

\begin{figure}[ht]
  \begin{center}
  {\bf Abalone $n = 2835$ \vspace{0.2cm}}
     \begin{tabular}{ccc}
      \hspace{-0.4cm} \includegraphics[width=\halffigwidth]{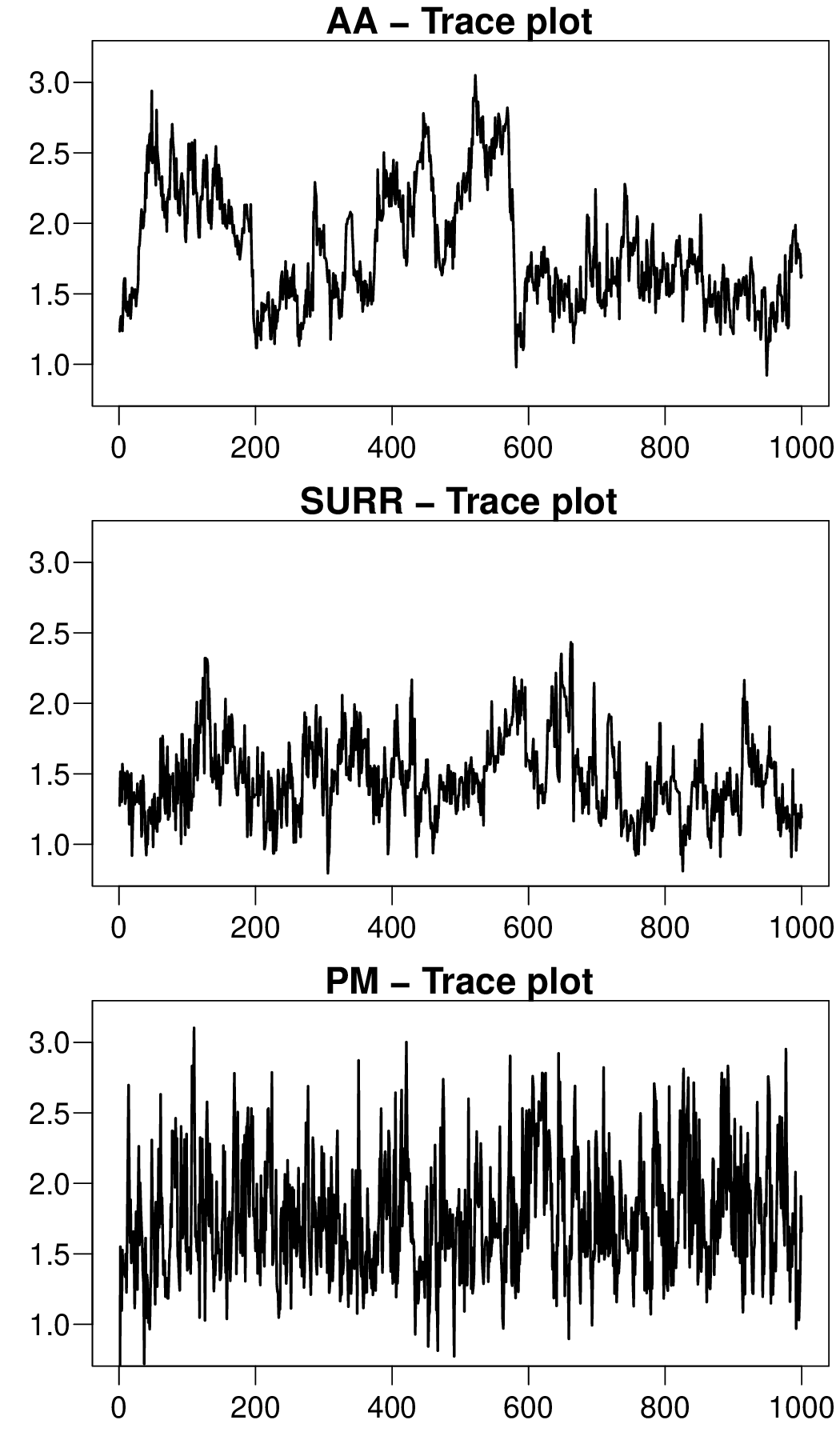}  & \hspace{-0.5cm} \includegraphics[width=\quarterfigwidth]{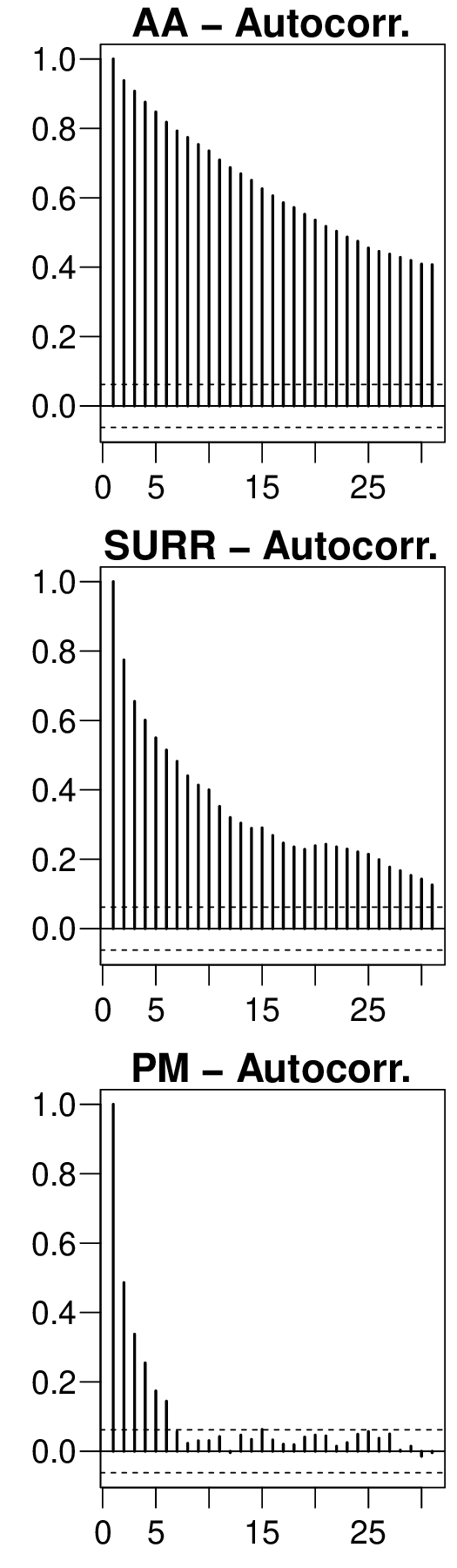} & \hspace{-0.5cm} \includegraphics[width=\quarterfigwidth]{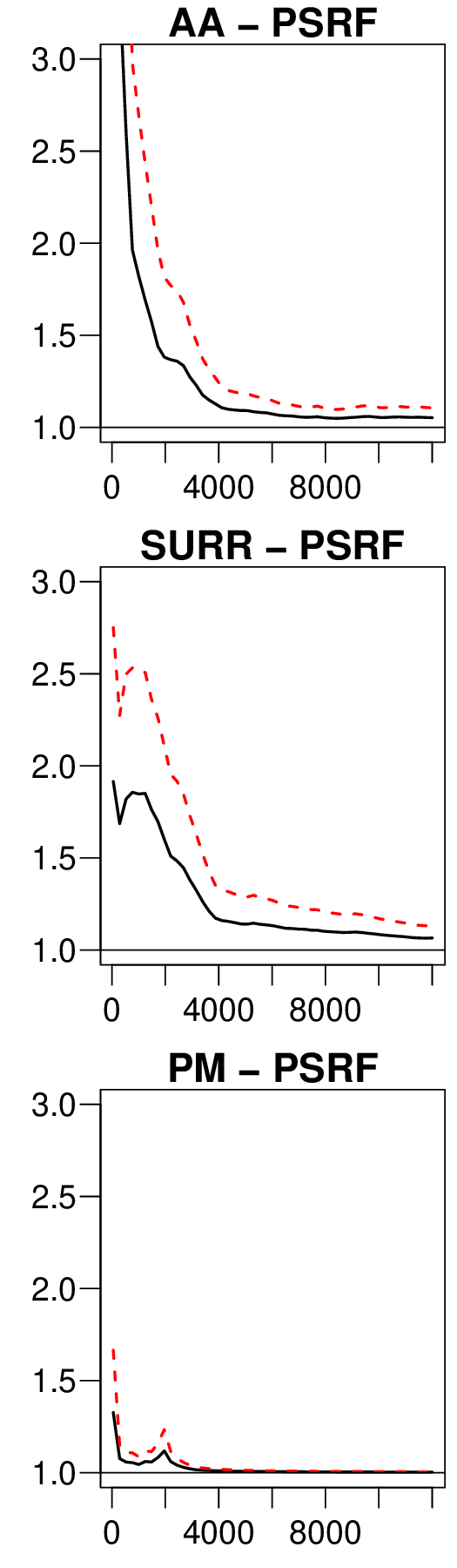} \\
    \end{tabular}
  \end{center}
  \caption{Summary of efficiency and convergence speed on the Abalone data set.}  \label{fig:sampling:abalone}
\end{figure}

The results indicate that the SURR parameterization yields better performance compared to the AA parameterization, where convergence can be extremely slow.
The PM approach achieves impressive efficiency and convergence speed compared to the AA and SURR parameterizations.
Remarkably, in these experiments, this is achieved using only one importance sample and a relatively cheap approximation based on the LA algorithm. 
We note here that this might not be the case in general.
The key to the success of the PM approach is the possibility to obtain a low-variance estimator for the marginal likelihood $p(\yvect | \thetavect)$.
Some approximations might not be good enough to ensure that this is the case; in such situations, more importance samples or better approximations should be employed.
This has recently been investigated in more detail in~\cite{FilipponeEA_MCMC_AIS13}, where unbiased estimates of the marginal likelihood based on Annealed Importance Sampling~\cite{Neal01} have been proposed.

\section{Comparison with other probabilistic classifiers}

\subsection{Data sets and experimental setup}
This section reports a comparison of classification performance on five UCI data sets.
In the Glass data set we considered the two classes ``window glass'' and ``non-window glass''.
In the USPS data set we considered the task of classification of ``3'' vs ``5'' as in~\cite{Kuss05}.

We constructed increasingly larger training sets comprising an equal number of input vectors per class.
For each value of number of input vectors $n$, we constructed $40$ training sets across which we evaluated the performance of the proposed PM approach.

\subsection{Comparing methods}
We compared the PM approach (MCMC PM EP) with (i) a probabilistic version of an SVM classifier~\cite{Chang11} (ii) the GP classifier using the EP approximation~\cite{Kuss05,Rasmussen06} optimizing $\thetavect$ using type II Maximum Likelihood~\cite{Rasmussen06,Bishop06} (EP ML) and (iii) with the classifier obtained by sampling $\thetavect$ based on the marginal likelihood computed by EP (MCMC EP).
Predictions in EP ML and MCMC EP were carried out by approximately integrating out latent variables according to the Gaussian approximation given by the EP algorithm.

We used the SVM code in the \texttt{e1071} R package, which provides a front-end to the \texttt{LIBSVM} library where non-linear SVMs employ a squared exponential isotropic kernel.
In order to meaningfully compare the four classifiers, we used the isotropic kernel/covariance for all of them.   \label{REV:1:7} 
In the MCMC PM EP method we set $N_{\mathrm{imp}} = 64$.

\subsection{Performance scores}
We are interested in comparing the ability of the classifiers to reliably quantify uncertainty in predicting class labels.   \label{REV:1:9}
Predictive probabilities give a means to do so; the more confident the classifier is about a correct class label, the better.
Also, predictive probabilities make it possible to avoid making decisions when predictive probabilities are below a given threshold.

Following~\cite{Ferri04}, we propose to summarize the ability of a classifier to reliably quantify uncertainty in predictions by analyzing classification accuracy and AUC (which denotes the area under the Receiver Operating Characteristic (ROC) curve for the classifier) versus the degree of abstention.
In particular, we propose to measure the area under the two curves obtained by plotting accuracy versus degree of abstention and AUC versus degree of abstention.
We will denote such scores by ``capacity accuracy'' and ``capacity AUC'', respectively.
A value of capacity close to one suggests that a classifier is capable of correctly classifying test data with a high degree of confidence.

For a probabilistic classifier, we compute accuracy and AUC versus degree of abstention as follows.
Denote by $\rho$ a threshold for the predictive probabilities.
According to a threshold value $\rho$, we compute the degree of abstention as the proportion of test data for which the predictive probability $p$ satisfies the following condition $0.5 - \rho < p < 0.5 + \rho$.
For the rest of the test data, namely data for which the classifier is most confident, we compute accuracy and AUC.
We repeat this procedure for different values of $\rho$, starting from $0.00$ going up to $0.50$ at increments of $0.01$, so that we obtain the plots of accuracy and AUC with respect to degree of abstention.
We finally compute the area of the two curves to obtain ``capacity accuracy'' and ``capacity AUC'' for the classifier.
Given that the degree of abstention might not reach the value of one, we propose to divide the area of the curves by the largest degree of abstention, so that the two capacity scores are normalized to one.
Figure~\ref{fig:results:vs:abstention} shows two exemplar curves of accuracy and AUC with respect to the degree of abstention that are used to compute the capacity scores.

\begin{figure}[ht]
  \begin{center}
     \begin{tabular}{cc}
      \hspace{-0.4cm} \includegraphics[width=\figwidth,height=\figwidth]{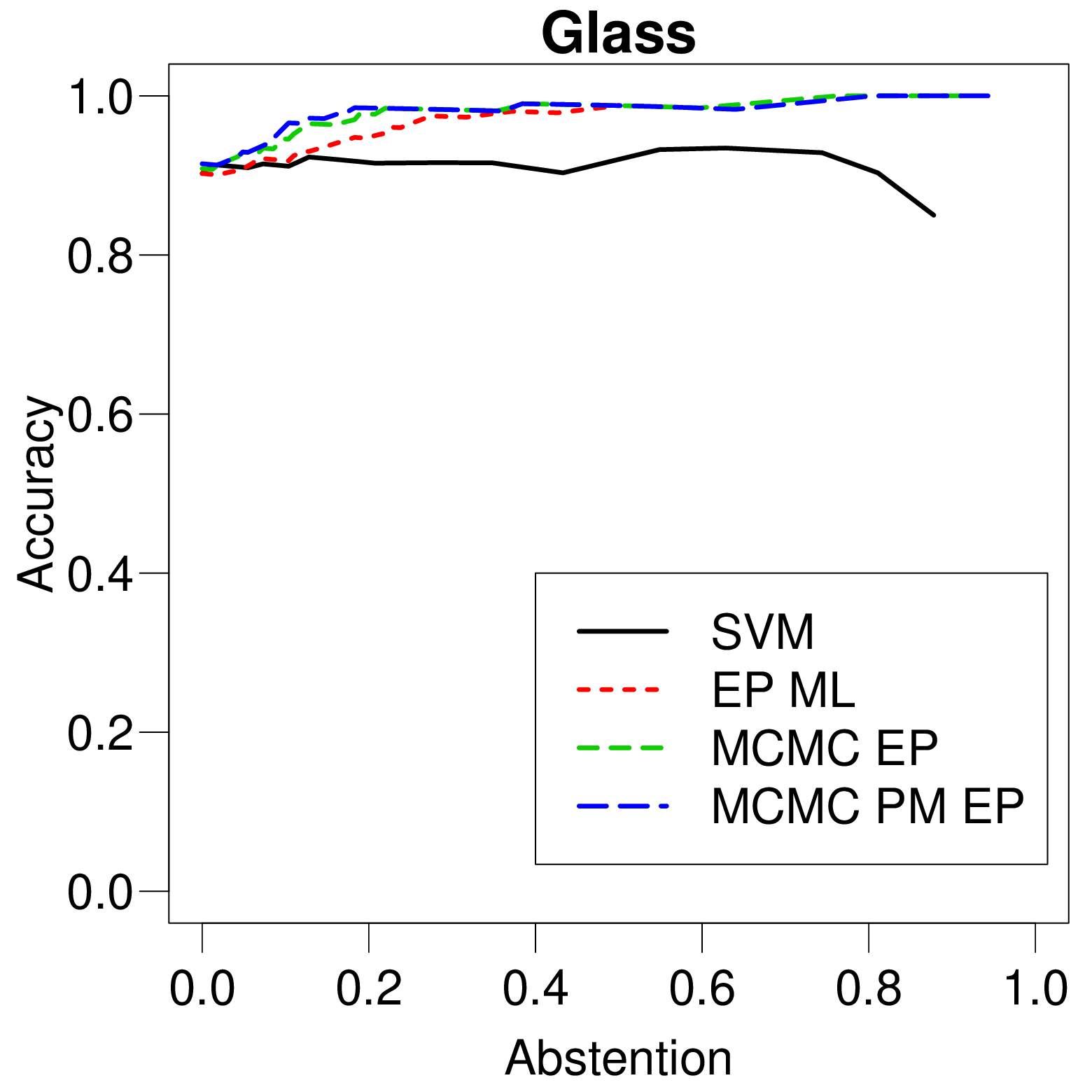}  & \hspace{-0.5cm} \includegraphics[width=\figwidth,height=\figwidth]{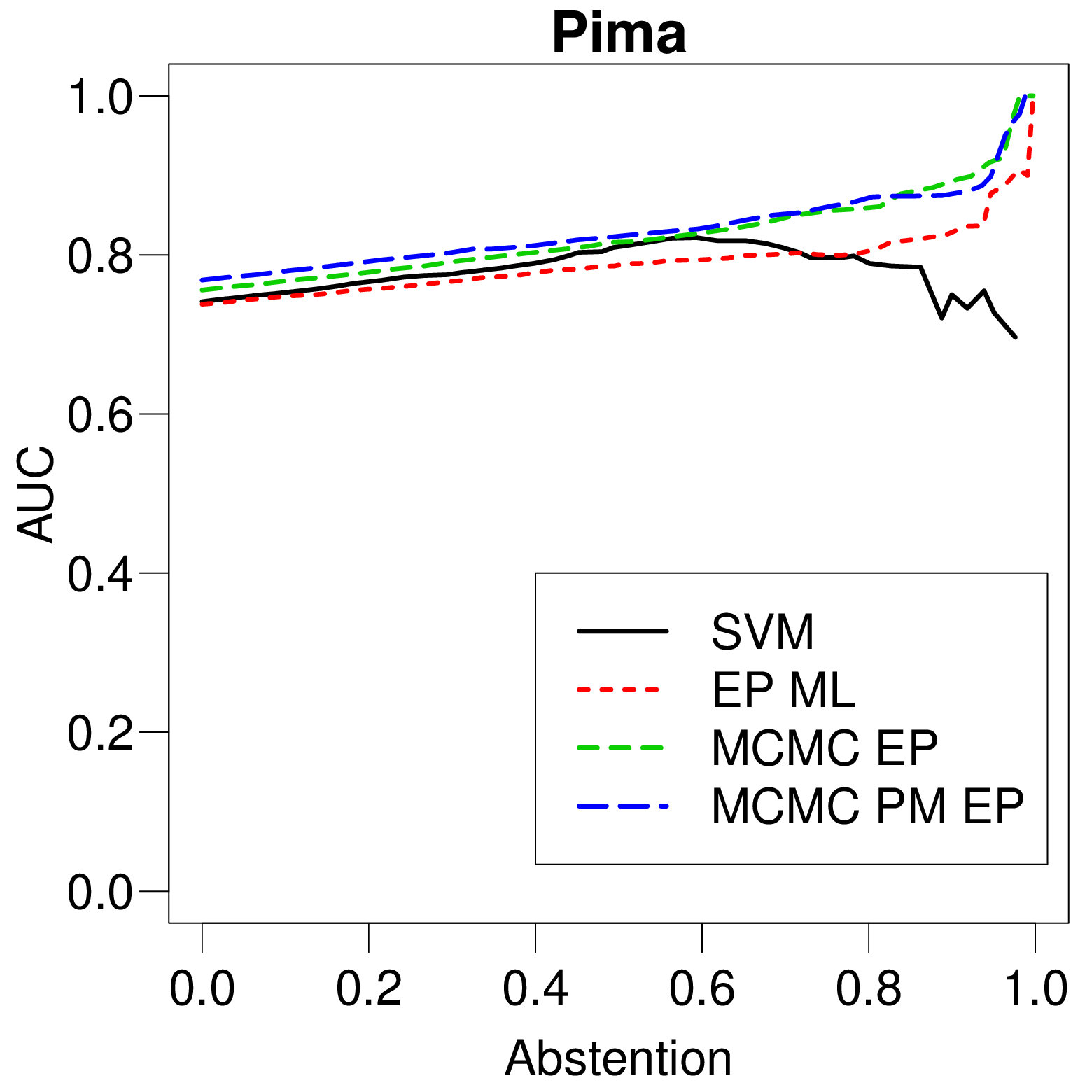} \\
    \end{tabular}
  \end{center}
  \caption{Left panel: Plot of accuracy vs degree of abstention for one of the folds in the Glass data set for $n=50$. Right panel: Plot of AUC vs degree of abstention for one of the folds in the Pima data set for $n=20$.}  \label{fig:results:vs:abstention}
\end{figure}

\subsection{Results}
The results are reported in figures \ref{fig:results:real1} and \ref{fig:results:real2} for the five data sets considered in this work.
In general, the probabilistic version of the SVM classifier leads to a worse quantification of uncertainty compared to the GP classifiers.
Figure~\ref{fig:results:vs:abstention} shows how SVMs tend to assign high confidence to wrong decisions more often than GP classifiers.
Also, the GP classifier optimizing the hyper-parameters (EP~ML) yields worse performance compared to the GP classifiers where hyper-parameters are integrated out.
Finally, the general trend is that the PM approach is the one achieving the best quantification of uncertainty compared to all the classifiers considered in this work. 

\begin{figure}[ht]
  \begin{center}
     \begin{tabular}{cc}
      \hspace{-0.4cm} \includegraphics[width=\figwidth,height=\figwidth]{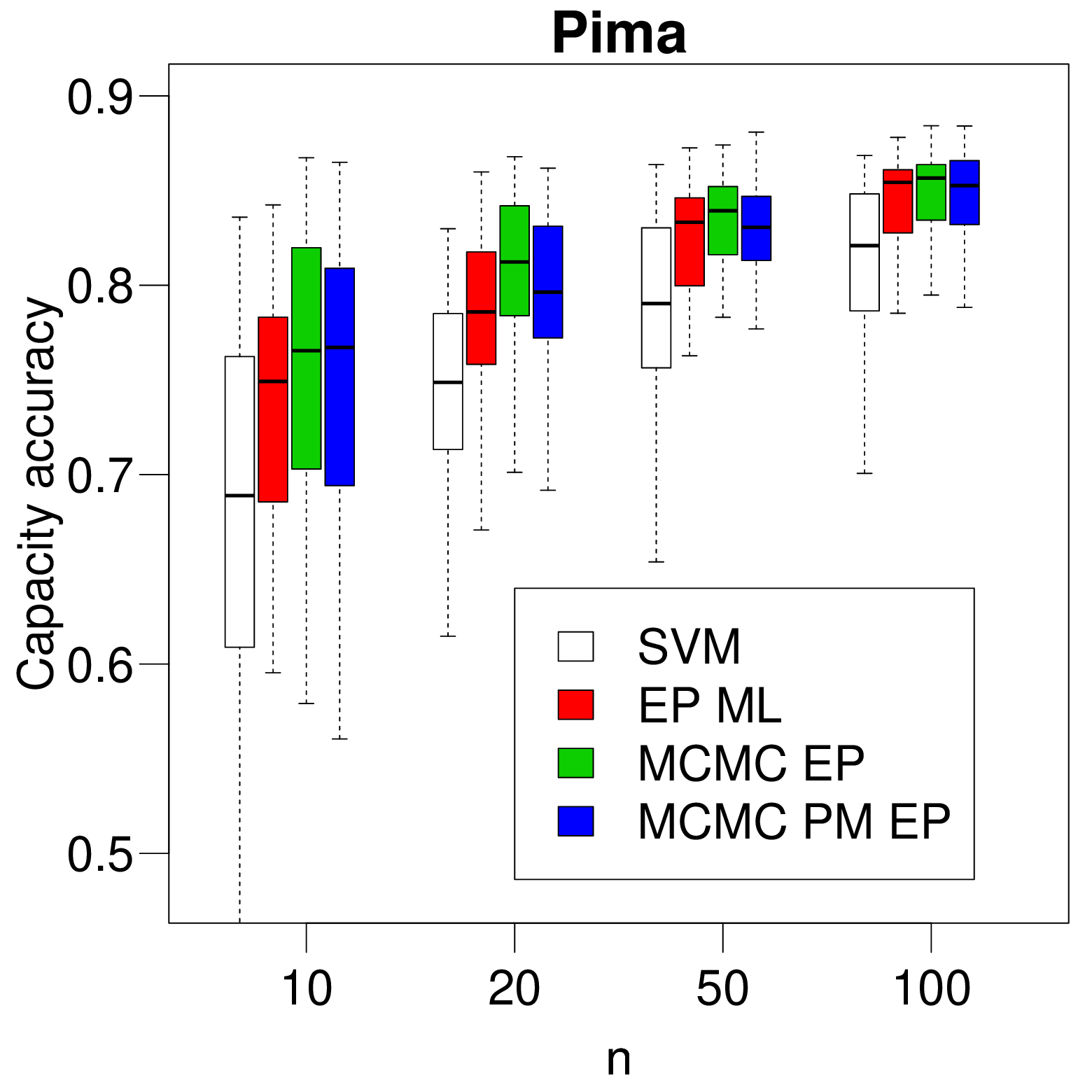}  & \hspace{-0.5cm} \includegraphics[width=\figwidth,height=\figwidth]{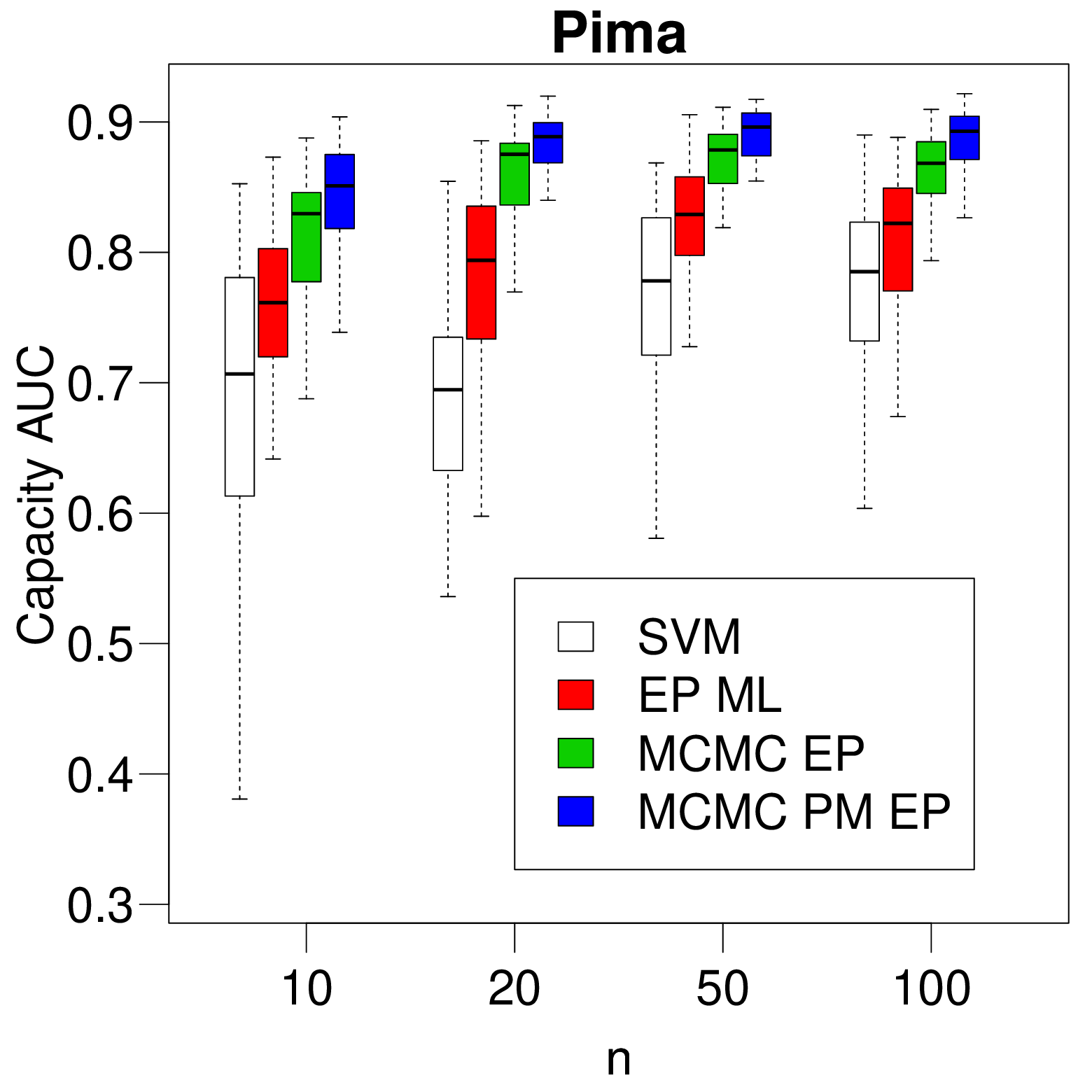} \\
      \hspace{-0.4cm} \vspace{-0.4cm}  \includegraphics[width=\figwidth,height=\figwidth]{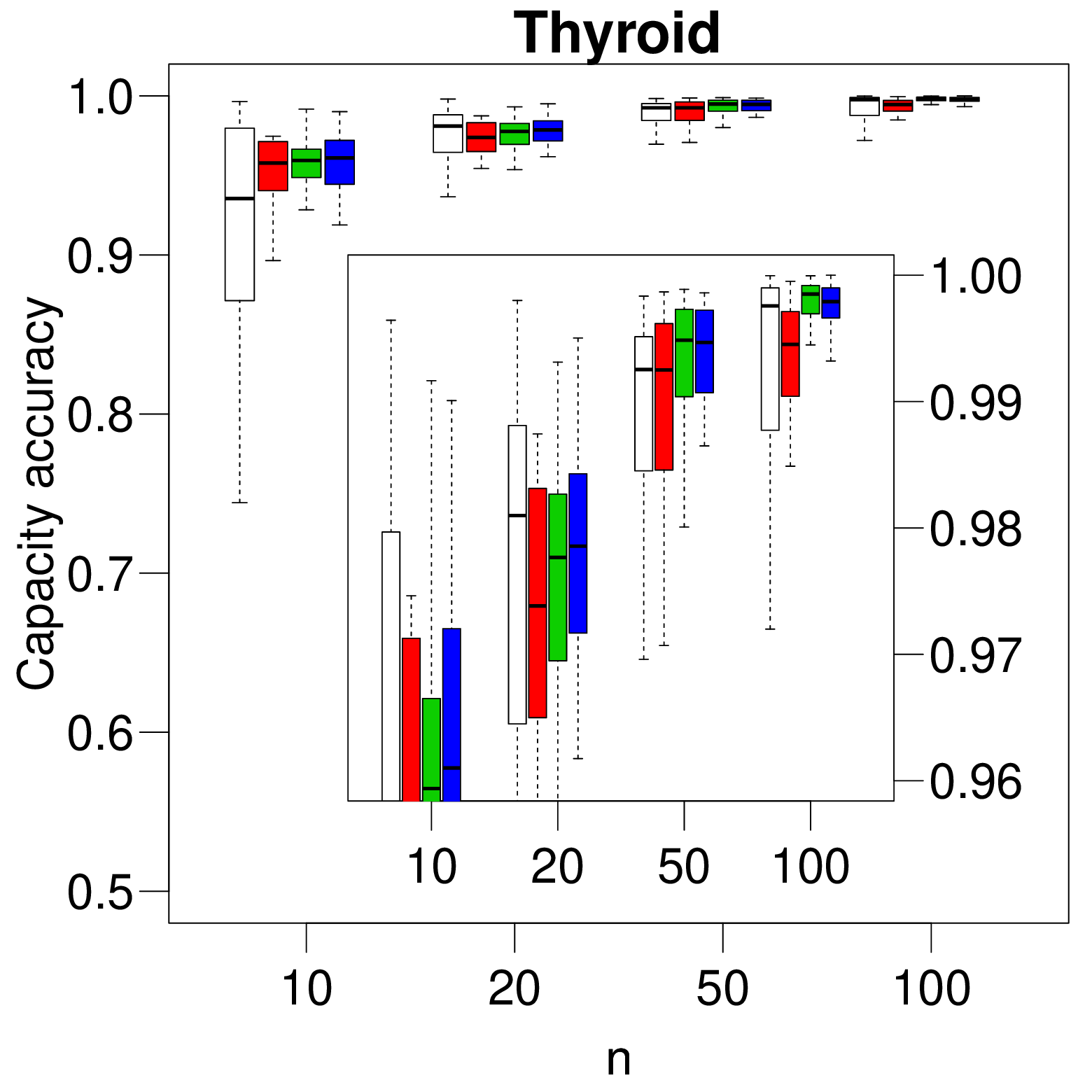}  & \hspace{-0.5cm} \includegraphics[width=\figwidth,height=\figwidth]{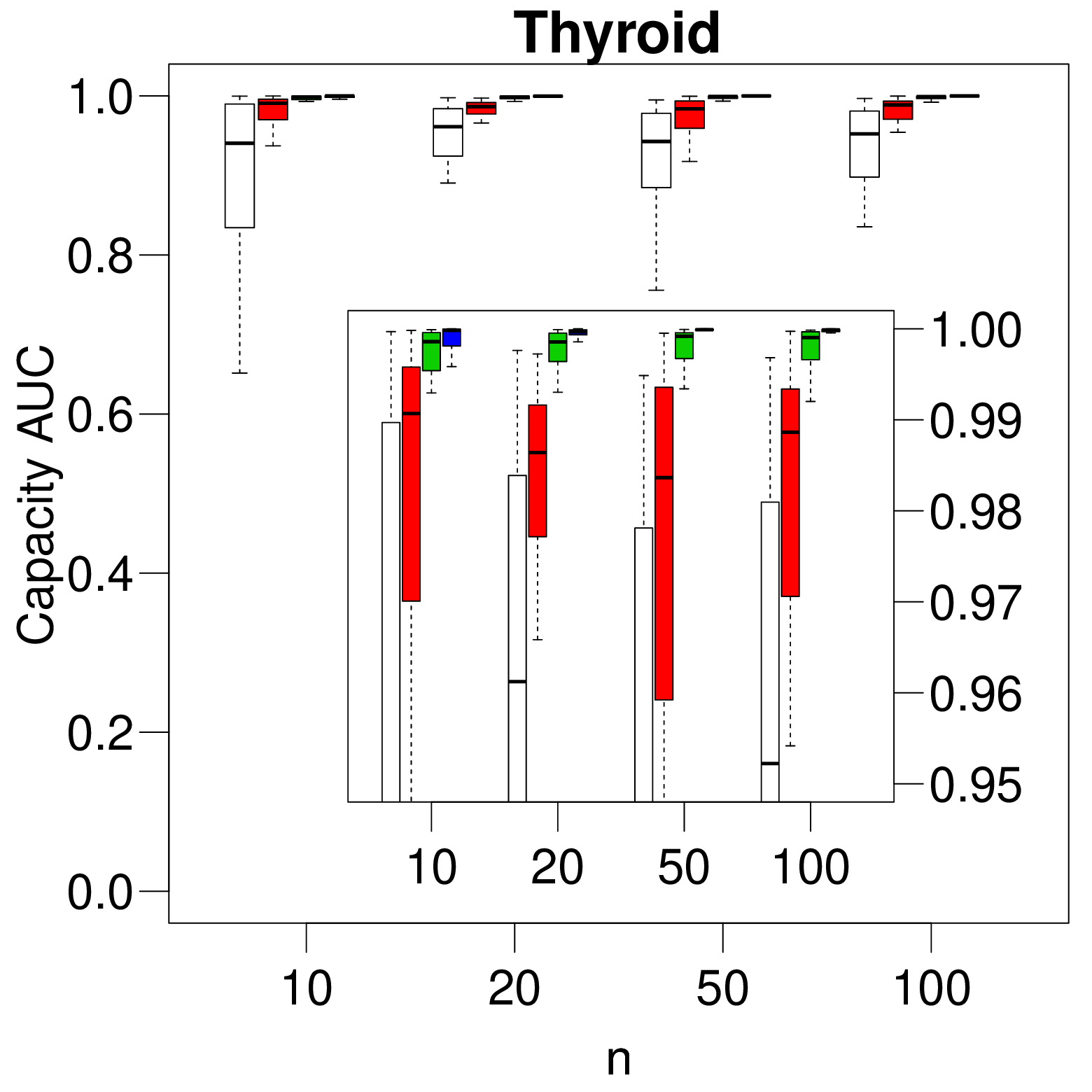} \\
    \end{tabular}
  \end{center}
  \caption{Plots of performance scores with respect to size of training set for the Pima (first row) and the Thyroid (second row) data sets. The legend is reported in the first row only and it applies to all the plots. In the remaining plots, a closeup is reported to make it easier to compare the results.}  \label{fig:results:real1}
\end{figure}

\begin{figure}[ht]
  \begin{center}
    \begin{tabular}{cc}
      \hspace{-0.4cm} \includegraphics[width=\figwidth,height=\figwidth]{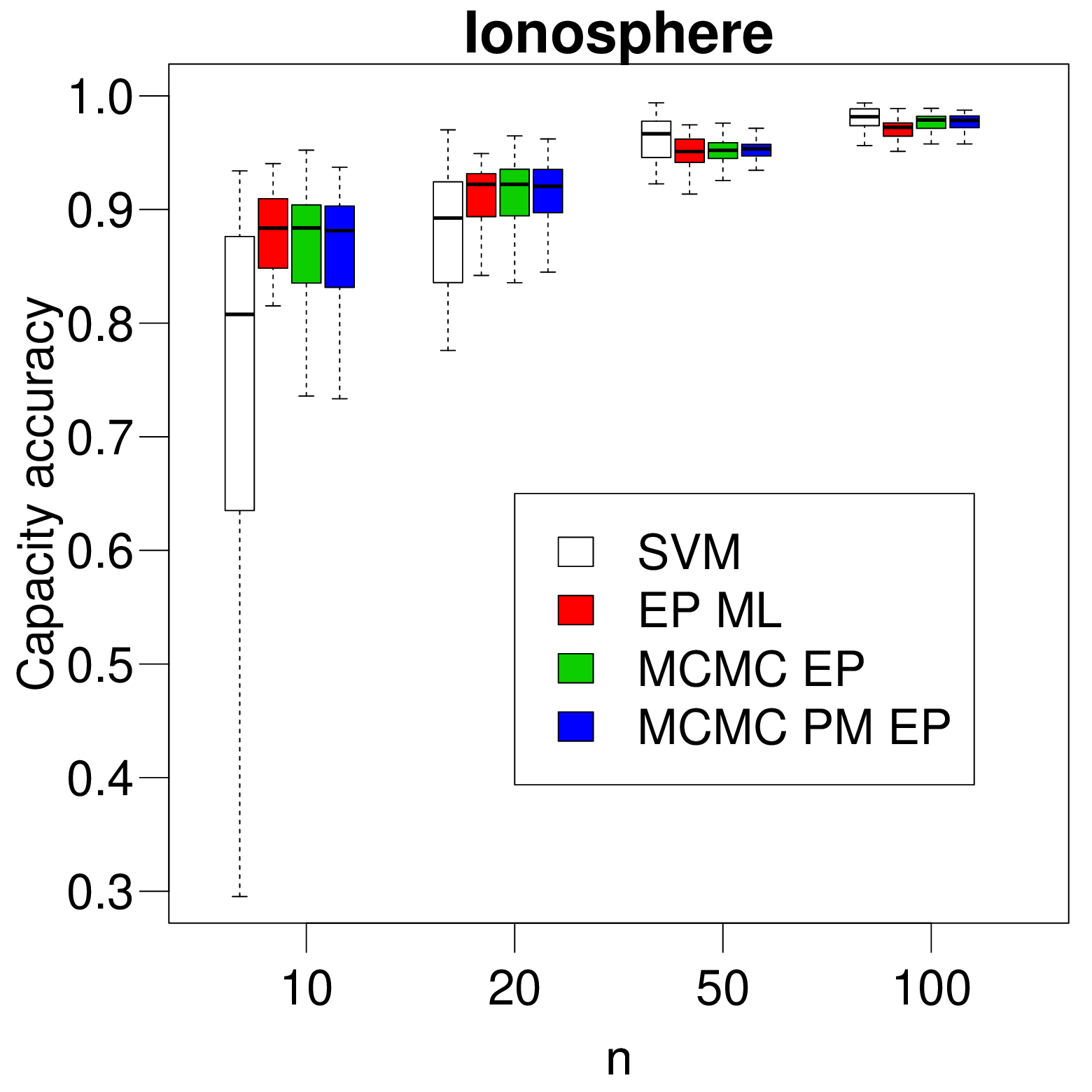}  & \hspace{-0.5cm} \includegraphics[width=\figwidth,height=\figwidth]{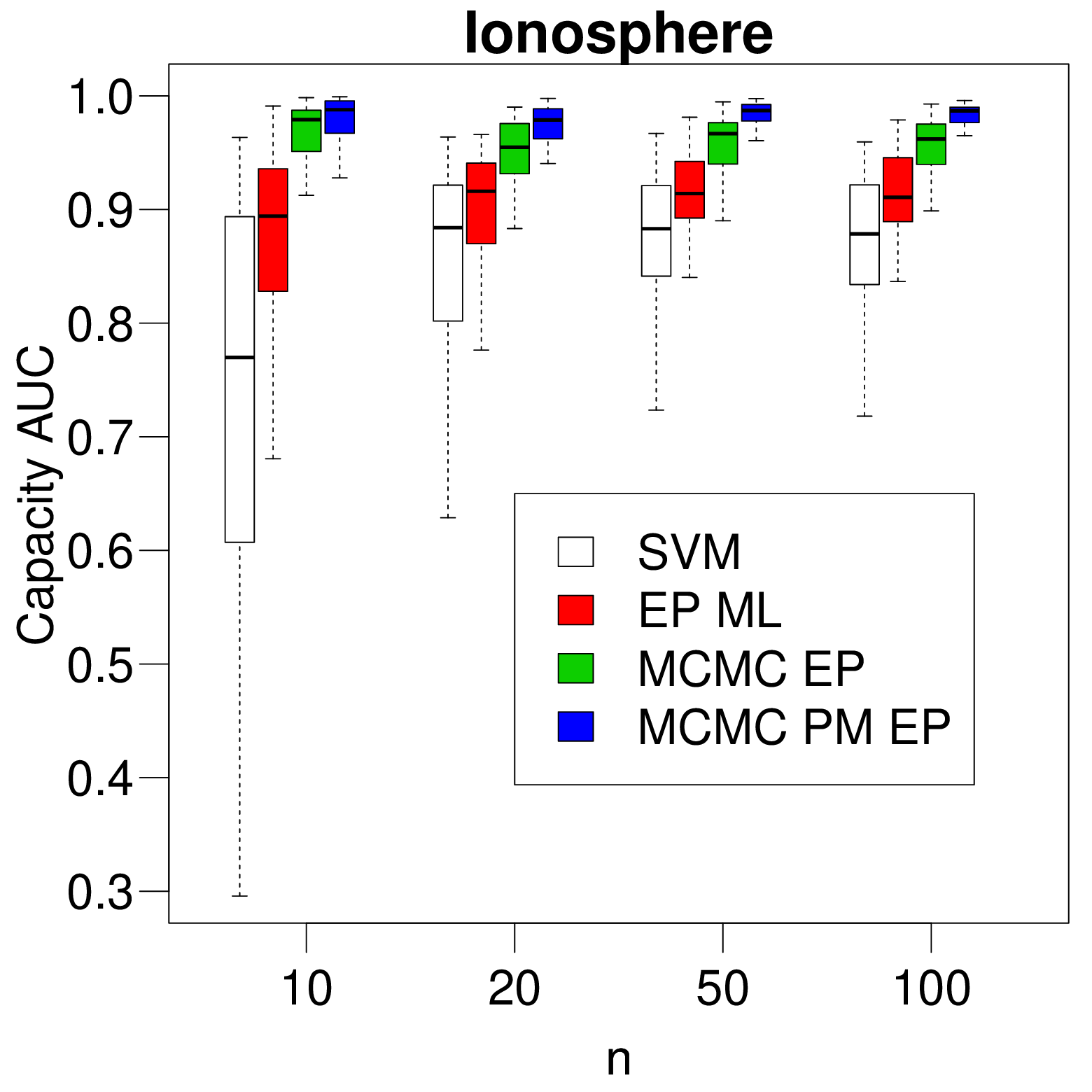} \\
      \hspace{-0.4cm} \includegraphics[width=\figwidth,height=\figwidth]{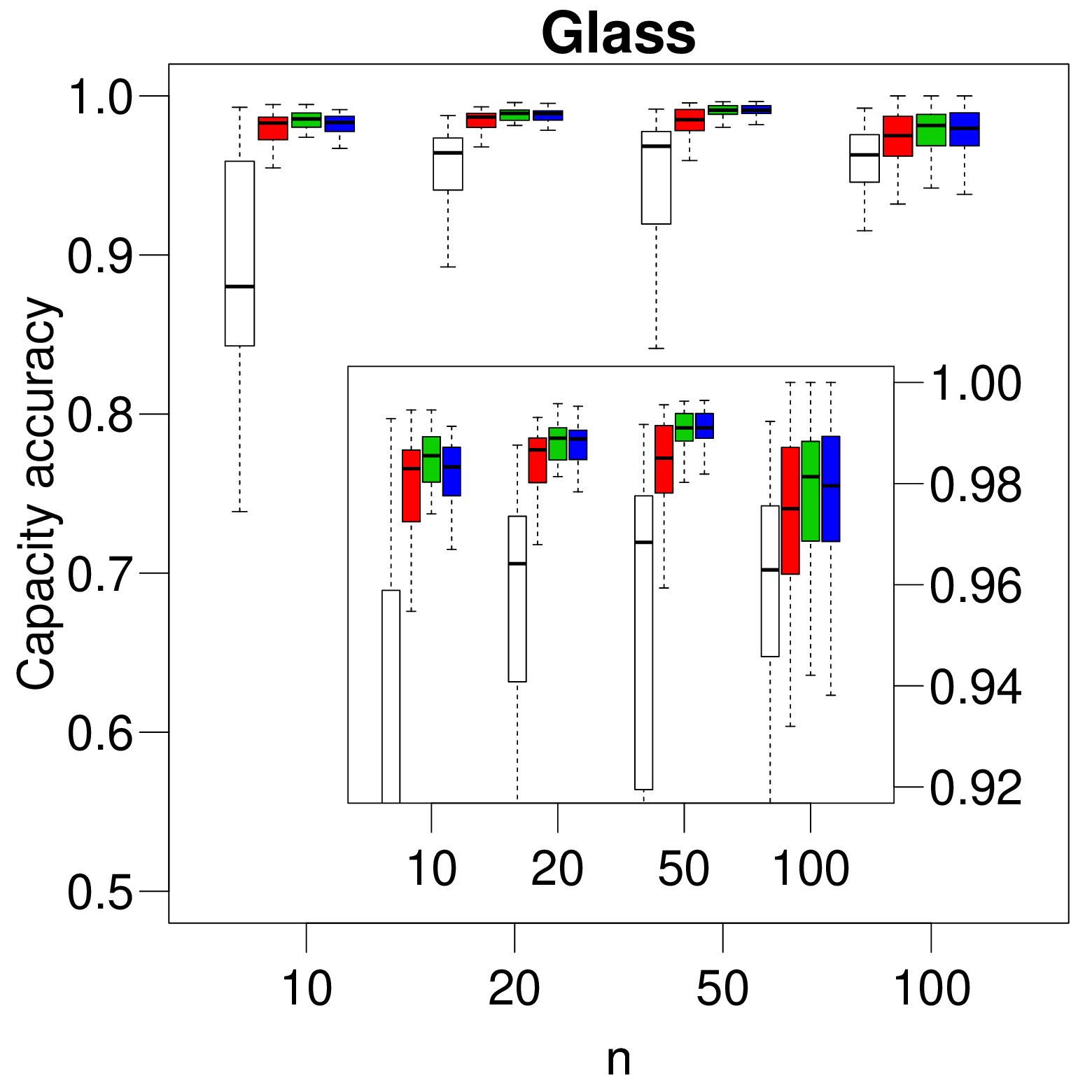}  & \hspace{-0.5cm} \includegraphics[width=\figwidth,height=\figwidth]{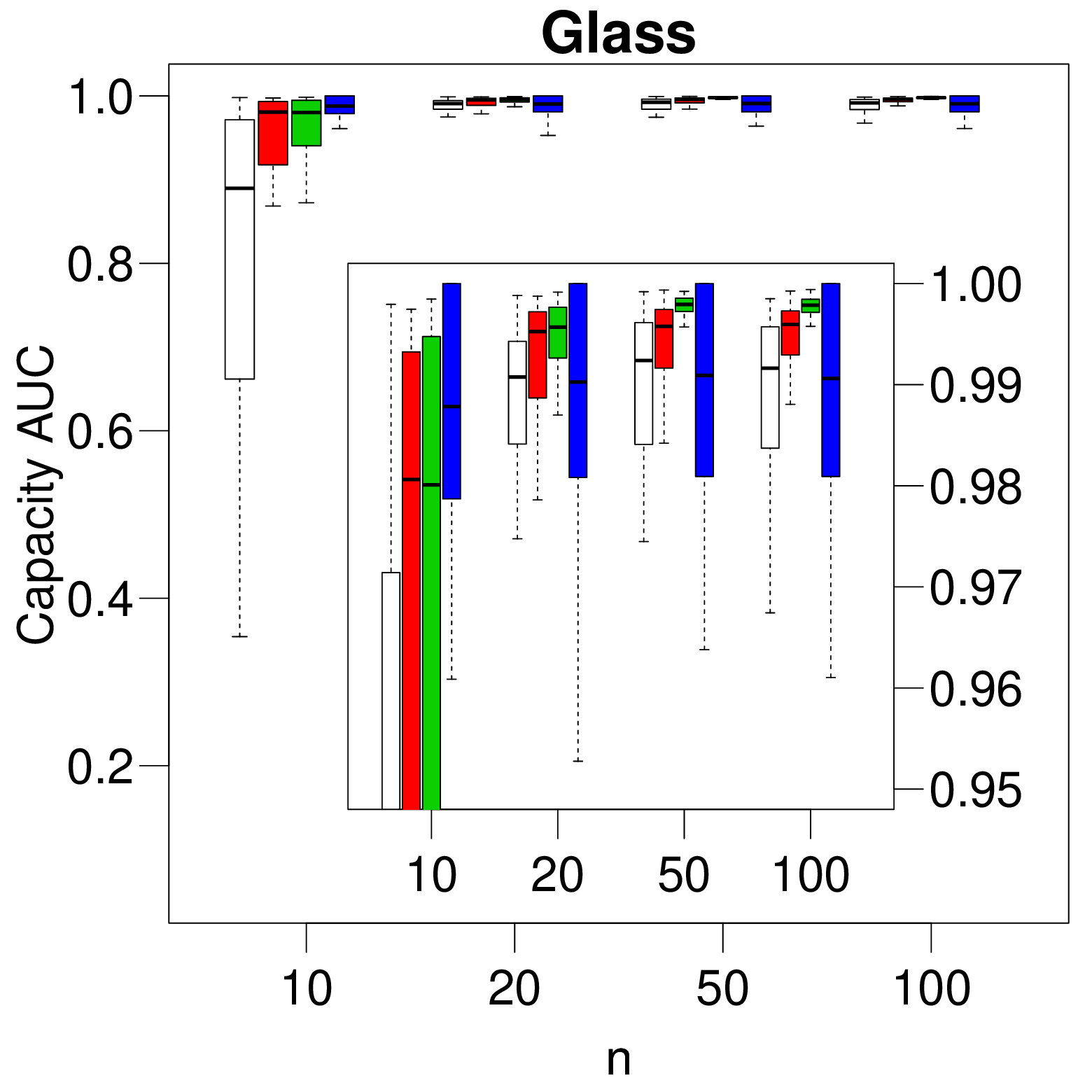} \\
      \hspace{-0.4cm} \vspace{-0.4cm}  \includegraphics[width=\figwidth,height=\figwidth]{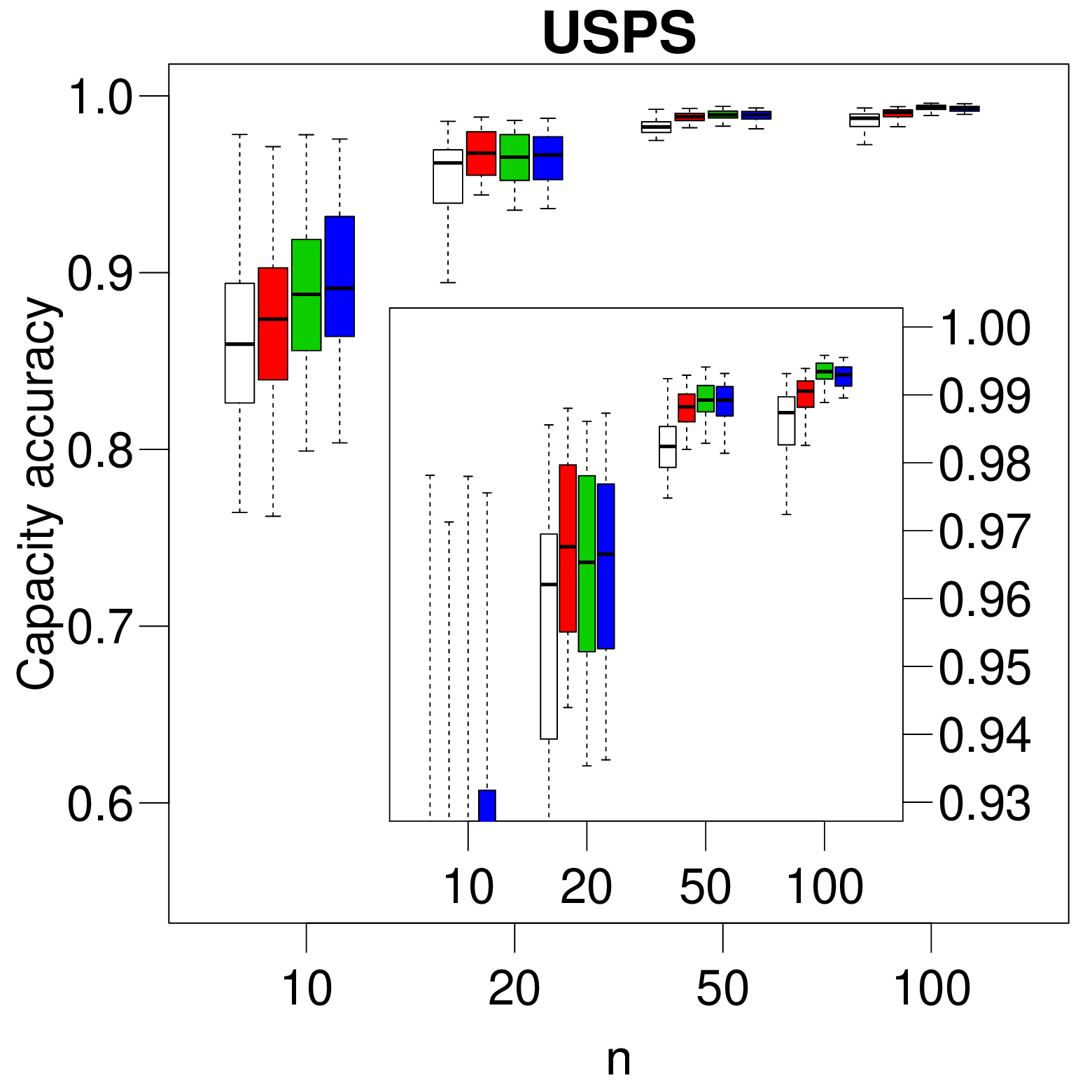}  & \hspace{-0.5cm} \includegraphics[width=\figwidth,height=\figwidth]{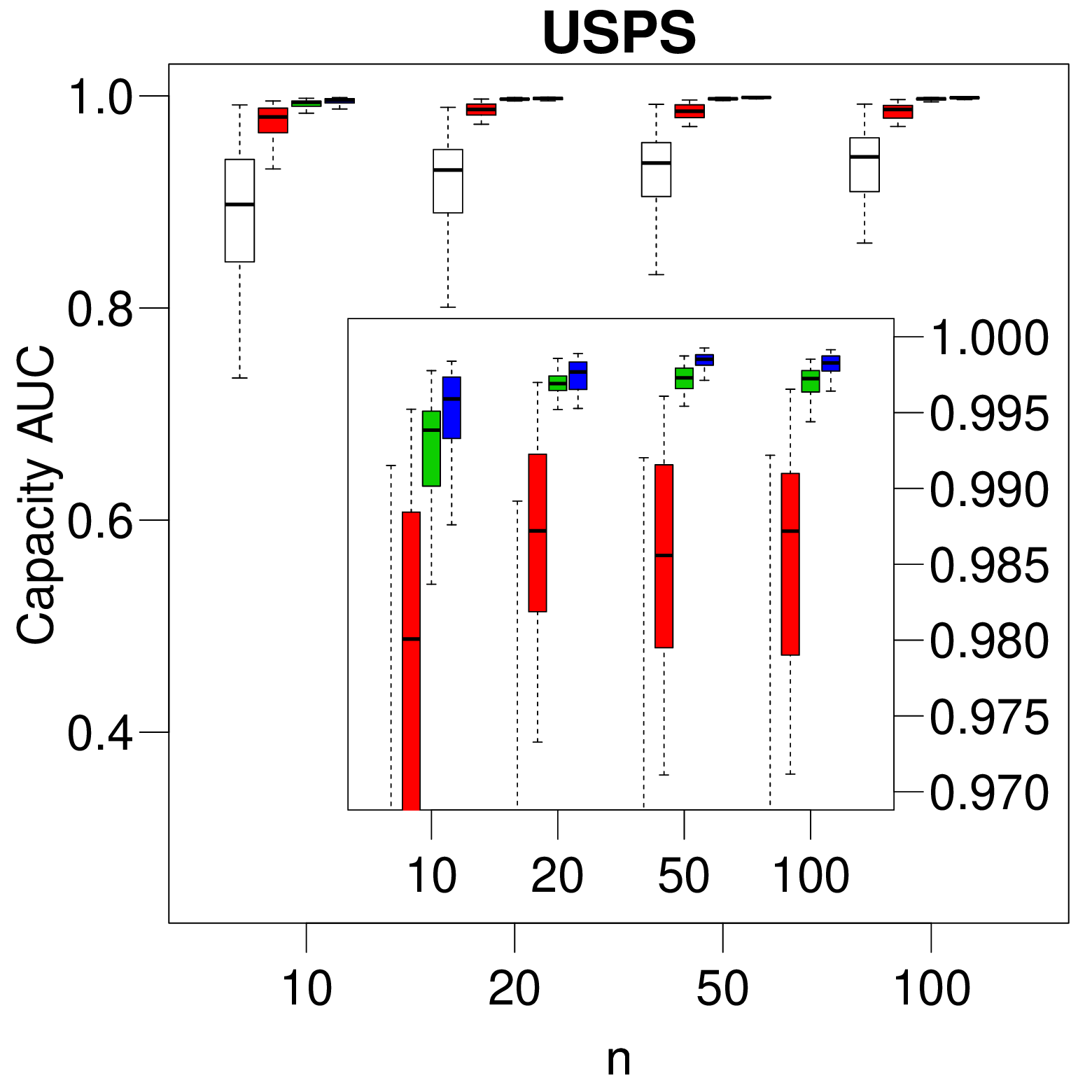} \\
    \end{tabular}
  \end{center}
  \caption{Plots of performance scores with respect to size of training set for the Ionosphere (first row), the Glass (second row) and the USPS (third row) data sets. The legend is reported in the first row only and it applies to all the plots. In the remaining plots, a closeup is reported to make it easier to compare the results.}  \label{fig:results:real2}
\end{figure}

A closer look at the posterior distribution obtained by MCMC~EP and MCMC~PM~EP reveals the following insights. \label{REV:1:10}
In the case of classification, the quality of the approximation of the marginal likelihood given by EP is generally accurate enough that the posterior distribution over the hyper-parameters for the MCMC~PM~EP approach is similar to the one obtained by employing MCMC~EP.
Differences in predictions obtained by the two methods are mostly due to the different way latent variables are integrated out when making predictions.
The situation can be different for other GP models where, for example, EP cannot be derived or EP exhibits convergence issues, so those considerations are peculiar to GP classification.
The aim of this study is to demonstrate that 
 it is important to account for the uncertainty in the hyper-parameters when predicting labels on unseen data in order to achieve a sound quantification of uncertainty.
Although this has been pointed out in previous work~\cite{Bishop06,Neal99,Rue09,Taylor12}, the proposed approach makes exact Bayesian computations, in a Monte Carlo sense, actually feasible by building upon deterministic approximations.

In terms of complexity of the three GP classifiers, the following considerations can be made. \label{REV:1:8} 
All the methods employ EP that requires three $O(n^3)$ operations at each iteration.
In the case of ML, the approximation of the marginal likelihood $p(\yvect | \thetavect)$ given by EP is optimized with respect to $\thetavect$. 
When the number of hyper-parameters is small, as in the cases of the isotropic RBF covariance function considered here, the optimization can be performed by grid search.
In the case of the ARD covariance, optimization can be performed, for instance, by employing the conjugate gradient algorithm, that requires the derivatives of the approximate marginal likelihood with respect to the hyper-parameters.
Computing such derivatives involves an extra $O(n^3)$ operation (see section 5.5.2 of~\cite{Rasmussen06}).
In MCMC~EP the approximation of the marginal likelihood $p(\yvect | \thetavect)$ given by EP is used directly to obtain samples form $p(\thetavect | \yvect)$.
In this case, each iteration requires running EP to obtain an approximation to $p(\yvect | \thetavect)$, so the overall complexity is still in $O(n^3)$, but the number of operations depends on the number of the iterations the MCMC approach is run for.
Running MCMC~PM~EP requires exactly the same number of operations as MCMC~EP, except that the Cholesky decomposition of the covariance of the approximating Gaussian is needed to draw the importance samples, adding an extra $O(n^3)$ operation.

To compute the mean of the predictive distribution for one test sample, EP~ML requires only operations in $O(n^2)$ and none in $O(n^3)$, as the expensive elements needed to compute it are already available from running the EP approximation.
In the case of MCMC~EP, the mean of the predictive distribution is an average of means computed for several draws from the posterior over $\thetavect$, so the complexity scales linearly with the number of MCMC samples and quadratically with $n$.
MCMC~PM~EP, instead, requires samples from the posterior distribution over latent variables in addition to hyper-parameters.
Drawing samples from $p(\fvect | \yvect, \thetavect)$ requires computations in $O(n^2)$ as previously discussed.
For each sample from the posterior distribution over latent variables and hyper-parameters, all the other computations are again in $O(n^2)$ following similar arguments as in the case of EP~ML; therefore, similarly to MCMC~EP, the complexity scales linearly with the number of MCMC samples and quadratically with $n$.

\section{Conclusions}

This paper presented a methodology that enables the fully Bayesian treatment of GP models, using probit regression as a working example, and builds upon existing approximate methods for integrating out latent variables.
The key element in this paper is the adoption of the pseudo marginal approach to devise a correct MCMC sampling scheme for the hyper-parameters of the covariance of the Gaussian Process prior from an approximation of the marginal density of the observation given the hyper-parameters.
The resulting sampling scheme is simple, and it is based on approximate methods that are currently very popular.

The results indicate that the proposed methodology leads to an MCMC approach where chains are characterized by high convergence speed and high efficiency.
This is an important feature that yields a step forward toward making fully Bayesian inference based on Gaussian Processes a concrete possibility for many applications with little user intervention.
The overall efficiency is driven by the MH proposal for the hyper-parameters that can be inefficient for models with several hyper-parameters; a matter of current investigation is to study alternative proposal mechanisms that avoid the erratic behavior of random walk exploration.

In support vector based classifiers hyper-parameters are optimized by minimizing the cross-validation error across a set of candidate values.
It is clear that for small data sets or for covariance functions with a large number of hyper-parameters, this procedure becomes unfeasible.
The proposed approach, instead, yields a natural way to integrate the uncertainty in the hyper-parameters when making predictions and infer them from data.
The comparison with other state-of-the-art probabilistic classifiers that are commonly employed in the Machine Learning community shows that accounting for the posterior over the hyper-parameters is extremely beneficial, especially for small data sets.

In terms of scalability, the main computational bottleneck is in the computation of the GP prior density that requires the factorization of the covariance matrix, which is in $O(n^3)$.
The same considerations apply to all GP classifiers that use approximate methods to integrate out latent variables, so we argue that by running an efficient sampling procedure for the hyper-parameters rather than an optimization strategy, the computational overhead will not be dramatically higher, but the classification performance will be much more reliable.

We believe that the results presented here can be extended to other latent Gaussian models, such as Log-Gaussian Cox process models~\cite{Moller98} and Ordinal Regression with GP priors~\cite{Chu05}.
Finally, it is possible to extend the proposed PM MCMC approach to deal with GP models characterized by a sparse inverse covariance, which are popular when analyzing spatio-temporal data.
In this case, it is possible to exploit sparsity in the inverse covariance of the GP, yielding a fast mixing and efficient MCMC approach capable of processing large amounts of data.


%



\ifCLASSOPTIONcompsoc
  \section*{Acknowledgments}
\else
  \section*{Acknowledgment}
\fi

The authors would like to thank the anonymous reviewers for their critical and constructive comments and suggestions.
Mark Girolami is supported by an EPSRC Established Career Research Fellowship EP/J016934/2, a Royal Society Wolfson Research Merit Award, and EPSRC Project grants ENGAGE EP/K015664/2 and EQUIP EP/K034154/1.
This work is dedicated to \mbox{Stefano\ Filippone}.

\ifCLASSOPTIONcaptionsoff
  \newpage
\fi

\begin{IEEEbiography}[{\includegraphics[width=1in,height=1.25in,clip,keepaspectratio]{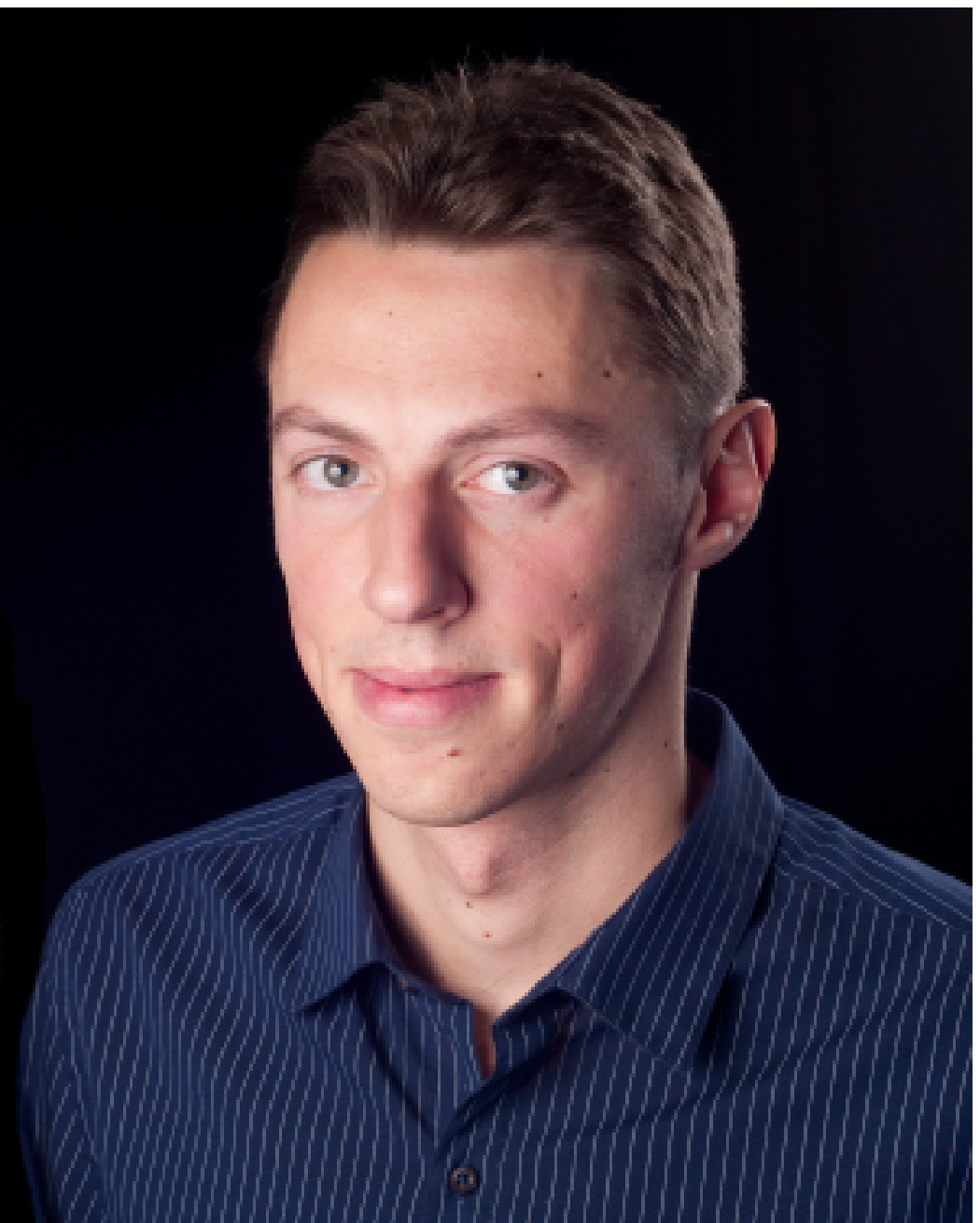}}]{Maurizio Filippone}

Maurizio Filippone received a Master's degree in Physics and a Ph.D. in Computer Science from the University of Genova, Italy, in 2004 and 2008, respectively. 

In 2007, he was a Research Scholar with George Mason University, Fairfax, VA.
From 2008 to 2011, he was a Research Associate with the University of Sheffield, U.K. (2008-2009), with the University of Glasgow, U.K. (2010), and with University College London, U.K (2011). 
He is currently a Lecturer with the University of Glasgow, U.K. 
His current research interests include statistical methods for pattern recognition.

Dr Filippone serves as an Associate Editor for Pattern Recognition and the IEEE Transactions on Neural Networks and Learning Systems.

\end{IEEEbiography}

\begin{IEEEbiography}[{\includegraphics[width=1in,height=1.25in,clip,keepaspectratio]{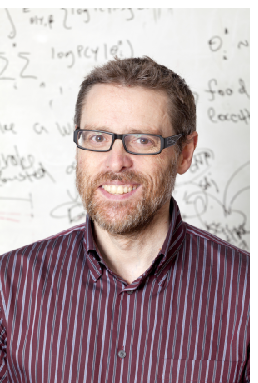}}]{Mark Girolami}
Mark Girolami FRSE is Professor of Statistics in the Department of
Statistics at the University of Warwick. He also holds a
professorial post in the Department of Computer Science at Warwick and is Director
of the EPSRC funded Research Network on Computational Statistics and Machine Learning. He is currently Editor-in-Chief of the journal
Statistics and Computing, an Associate Editor for J. R. Statist. Soc. C,
Journal of Computational and Graphical Statistics and Area Editor for Pattern
Recognition Letters. He currently holds a Royal Society Wolfson Research Merit
Award and an EPSRC Established Career Research Fellowship.
\end{IEEEbiography}


\begin{thebibliography}{10}
\providecommand{\url}[1]{#1}
\csname url@samestyle\endcsname
\providecommand{\newblock}{\relax}
\providecommand{\bibinfo}[2]{#2}
\providecommand{\BIBentrySTDinterwordspacing}{\spaceskip=0pt\relax}
\providecommand{\BIBentryALTinterwordstretchfactor}{4}
\providecommand{\BIBentryALTinterwordspacing}{\spaceskip=\fontdimen2\font plus
\BIBentryALTinterwordstretchfactor\fontdimen3\font minus
  \fontdimen4\font\relax}
\providecommand{\BIBforeignlanguage}[2]{{%
\expandafter\ifx\csname l@#1\endcsname\relax
\typeout{** WARNING: IEEEtran.bst: No hyphenation pattern has been}%
\typeout{** loaded for the language `#1'. Using the pattern for}%
\typeout{** the default language instead.}%
\else
\language=\csname l@#1\endcsname
\fi
#2}}
\providecommand{\BIBdecl}{\relax}
\BIBdecl

\bibitem{Cortes95}
C.~Cortes and V.~Vapnik, ``{Support Vector Networks},'' \emph{Machine
  Learning}, vol.~20, pp. 273--297, 1995.

\bibitem{Vapnik95}
V.~N. Vapnik, \emph{{The nature of statistical learning theory}}.\hskip 1em
  plus 0.5em minus 0.4em\relax New York, NY, USA: Springer-Verlag New York,
  Inc., 1995.

\bibitem{Rasmussen05}
C.~E. Rasmussen and J.~Q. Candela, ``{Healing the relevance vector machine
  through augmentation},'' in \emph{Proceedings of the 22nd international
  conference on Machine learning}, ser. ICML '05.\hskip 1em plus 0.5em minus
  0.4em\relax New York, NY, USA: ACM, 2005, pp. 689--696.

\bibitem{Tipping01}
M.~E. Tipping, ``{Sparse bayesian learning and the relevance vector machine},''
  \emph{Journal of Machine Learning Research}, vol.~1, pp. 211--244, 2001.

\bibitem{Rasmussen06}
C.~E. Rasmussen and C.~Williams, \emph{{Gaussian Processes for Machine
  Learning}}.\hskip 1em plus 0.5em minus 0.4em\relax MIT Press, 2006.

\bibitem{Bosch08}
A.~Bosch, A.~Zisserman, and X.~Muoz, ``{Scene Classification Using a Hybrid
  Generative/Discriminative Approach},'' \emph{IEEE Transactions on Pattern
  Analysis and Machine Intelligence}, vol.~30, no.~4, pp. 712--727, 2008.

\bibitem{FilipponeAOAS12}
M.~Filippone, A.~F. Marquand, C.~R.~V. Blain, S.~C.~R. Williams,
  J.~{Mour\~{a}o-Miranda}, and M.~Girolami, ``{Probabilistic Prediction of
  Neurological Disorders with a Statistical Assessment of Neuroimaging Data
  Modalities},'' \emph{Annals of Applied Statistics}, vol.~6, no.~4, pp.
  1883--1905, 2012.

\bibitem{Jaakkola00b}
T.~Jaakkola, M.~Diekhans, and D.~Haussler, ``{A Discriminative Framework for
  Detecting Remote Protein Homologies},'' \emph{Journal of Computational
  Biology}, vol.~7, no. 1-2, pp. 95--114, 2000.

\bibitem{Joachims98}
T.~Joachims, ``{Text Categorization with Suport Vector Machines: Learning with
  Many Relevant Features},'' in \emph{ECML}, ser. Lecture Notes in Computer
  Science, C.~Nedellec and C.~Rouveirol, Eds., vol. 1398.\hskip 1em plus 0.5em
  minus 0.4em\relax Springer, 1998, pp. 137--142.

\bibitem{Ratsch06}
G.~R{\"{a}}tsch, S.~Sonnenburg, and C.~Sch{\"{a}}fer, ``{Learning Interpretable
  SVMs for Biological Sequence Classification},'' \emph{BMC Bioinformatics},
  vol.~7, no. S-1, 2006.

\bibitem{OWilliams05}
O.~Williams, A.~Blake, and R.~Cipolla, ``{Sparse Bayesian learning for
  efficient visual tracking},'' \emph{IEEE Transactions on Pattern Analysis and
  Machine Intelligence}, vol.~27, no.~8, pp. 1292--1304, Aug. 2005.

\bibitem{Bishop06}
C.~M. Bishop, \emph{{Pattern Recognition and Machine Learning (Information
  Science and Statistics)}}, 1st~ed.\hskip 1em plus 0.5em minus 0.4em\relax
  Springer, Aug. 2007.

\bibitem{Williams98}
C.~K.~I. Williams and D.~Barber, ``{Bayesian classification with Gaussian
  processes},'' \emph{IEEE Transactions on Pattern Analysis and Machine
  Intelligence}, vol.~20, pp. 1342--1351, 1998.

\bibitem{Minka01}
T.~P. Minka, ``{Expectation Propagation for approximate Bayesian inference},''
  in \emph{Proceedings of the 17th Conference in Uncertainty in Artificial
  Intelligence}, ser. UAI '01.\hskip 1em plus 0.5em minus 0.4em\relax San
  Francisco, CA, USA: Morgan Kaufmann Publishers Inc., 2001, pp. 362--369.

\bibitem{Gibbs00}
M.~N. Gibbs and D.~J.~C. MacKay, ``{Variational Gaussian process
  classifiers},'' \emph{IEEE Transactions on Neural Networks and Learning
  Systems}, vol.~11, no.~6, pp. 1458--1464, 2000.

\bibitem{Rue09}
H.~Rue, S.~Martino, and N.~Chopin, ``{Approximate Bayesian inference for latent
  Gaussian models by using integrated nested Laplace approximations},''
  \emph{Journal of the Royal Statistical Society: Series B (Statistical
  Methodology)}, vol.~71, no.~2, pp. 319--392, 2009.

\bibitem{Opper00}
M.~Opper and O.~Winther, ``{Gaussian processes for classification: Mean-field
  algorithms},'' \emph{Neural Computation}, vol.~12, no.~11, pp. 2655--2684,
  2000.

\bibitem{Kuss05}
M.~Kuss and C.~E. Rasmussen, ``{Assessing Approximate Inference for Binary
  Gaussian Process Classification},'' \emph{Journal of Machine Learning
  Research}, vol.~6, pp. 1679--1704, 2005.

\bibitem{Nickisch08}
H.~Nickisch and C.~E. Rasmussen, ``{Approximations for Binary Gaussian Process
  Classification},'' \emph{Journal of Machine Learning Research}, vol.~9, pp.
  2035--2078, Oct. 2008.

\bibitem{Cseke11}
B.~Cseke and T.~Heskes, ``{Approximate Marginals in Latent Gaussian Models},''
  \emph{Journal of Machine Learning Research}, vol.~12, pp. 417--454, 2011.

\bibitem{FilipponeML13}
M.~Filippone, M.~Zhong, and M.~Girolami, ``A comparative evaluation of
  stochastic-based inference methods for {G}aussian process models,''
  \emph{Machine Learning}, vol.~93, no.~1, pp. 93--114, 2013.

\bibitem{Murray10}
I.~Murray and R.~P. Adams, ``{Slice sampling covariance hyperparameters of
  latent Gaussian models},'' in \emph{NIPS}, J.~D. Lafferty, C.~K.~I. Williams,
  J.~Shawe-Taylor, R.~S. Zemel, and A.~Culotta, Eds.\hskip 1em plus 0.5em minus
  0.4em\relax Curran Associates, 2010, pp. 1732--1740.

\bibitem{KnorrHeld02}
L.~Knorr-Held and H.~Rue, ``{On Block Updating in Markov Random Field Models
  for Disease Mapping},'' \emph{Scandinavian Journal of Statistics}, vol.~29,
  no.~4, pp. 597--614, Dec. 2002.

\bibitem{Andrieu09}
C.~Andrieu and G.~O. Roberts, ``{The pseudo-marginal approach for efficient
  Monte Carlo computations},'' \emph{The Annals of Statistics}, vol.~37, no.~2,
  pp. 697--725, Apr. 2009.

\bibitem{Beaumont03}
M.~A. Beaumont, ``{Estimation of Population Growth or Decline in Genetically
  Monitored Populations},'' \emph{Genetics}, vol. 164, no.~3, pp. 1139--1160,
  Jul. 2003.

\bibitem{Neal99}
R.~M. Neal, ``{Regression and classification using Gaussian process priors
  (with discussion)},'' \emph{Bayesian Statistics}, vol.~6, pp. 475--501, 1999.

\bibitem{Taylor12}
M.~B. Taylor and J.~P. Diggle, ``{INLA or MCMC? A Tutorial and Comparative
  Evaluation for Spatial Prediction in log-Gaussian Cox Processes},'' 2012,
  arXiv:1202.1738.

\bibitem{Chang11}
C.~C. Chang and C.~J. Lin, ``{LIBSVM}: A library for support vector machines,''
  \emph{ACM Transactions on Intelligent Systems and Technology}, vol.~2, no.~3,
  2011.

\bibitem{Mackay94}
D.~J.~C. Mac{k}ay, ``{B}ayesian methods for backpropagation networks,'' in
  \emph{Models of Neural Networks {III}}, E.~Domany, J.~L. van Hemmen, and
  K.~Schulten, Eds.\hskip 1em plus 0.5em minus 0.4em\relax Springer, 1994,
  ch.~6, pp. 211--254.

\bibitem{Tierney86}
L.~Tierney and J.~B. Kadane, ``{Accurate Approximations for Posterior Moments
  and Marginal Densities},'' \emph{Journal of the American Statistical
  Association}, vol.~81, no. 393, pp. 82--86, 1986.

\bibitem{Murray10b}
I.~Murray, R.~P. Adams, and D.~J.~C. MacKay, ``{Elliptical slice sampling},''
  \emph{Journal of Machine Learning Research - Proceedings Track}, vol.~9, pp.
  541--548, 2010.

\bibitem{Neal03}
R.~M. Neal, ``{Slice Sampling},'' \emph{Annals of Statistics}, vol.~31, pp.
  705--767, 2003.

\bibitem{Duane87}
S.~Duane, A.~D. Kennedy, B.~J. Pendleton, and D.~Roweth, ``{Hybrid Monte
  Carlo},'' \emph{Physics Letters B}, vol. 195, no.~2, pp. 216--222, 1987.

\bibitem{Neal93}
R.~M. Neal, ``Probabilistic inference using {M}arkov chain {M}onte {C}arlo
  methods,'' Dept. of Computer Science, University of Toronto, Tech. Rep.
  CRG-TR-93-1, Sep. 1993.

\bibitem{Girolami11}
M.~Girolami and B.~Calderhead, ``{Riemann manifold Langevin and Hamiltonian
  Monte Carlo methods},'' \emph{Journal of the Royal Statistical Society:
  Series B (Statistical Methodology)}, vol.~73, no.~2, pp. 123--214, Mar. 2011.

\bibitem{Yu11}
Y.~Yu and X.-L. Meng, ``{To Center or Not to Center: That Is Not the
  Question--An Ancillarity-Sufficiency Interweaving Strategy (ASIS) for
  Boosting MCMC Efficiency},'' \emph{Journal of Computational and Graphical
  Statistics}, vol.~20, no.~3, pp. 531--570, 2011.

\bibitem{Papaspiliopoulos07}
O.~Papaspiliopoulos, G.~O. Roberts, and M.~Sk\"{o}ld, ``{A general framework
  for the parametrization of hierarchical models},'' \emph{Statistical
  Science}, vol.~22, no.~1, pp. 59--73, 2007.

\bibitem{Robert05}
C.~P. Robert and G.~Casella, \emph{{Monte Carlo Statistical Methods (Springer
  Texts in Statistics)}}.\hskip 1em plus 0.5em minus 0.4em\relax Secaucus, NJ,
  USA: Springer-Verlag New York, Inc., 2005.

\bibitem{Friel08}
N.~Friel and A.~N. Pettitt, ``{Marginal likelihood estimation via power
  posteriors},'' \emph{Journal of the Royal Statistical Society: Series B
  (Statistical Methodology)}, vol.~70, no.~3, pp. 589--607, 2008.

\bibitem{Skilling06}
J.~Skilling, ``{Nested sampling for general Bayesian computation},''
  \emph{Bayesian Analysis}, vol.~1, no.~4, pp. 833--860, 2006.

\bibitem{Gelman92}
A.~Gelman and D.~B. Rubin, ``{Inference from iterative simulation using
  multiple sequences},'' \emph{Statistical Science}, vol.~7, no.~4, pp.
  457--472, 1992.

\bibitem{Geweke04}
J.~Geweke, ``{Getting it right: joint distribution tests of posterior
  simulators},'' \emph{Journal of the American Statistical Association},
  vol.~99, no. 467, pp. 799--804, 2004.

\bibitem{Gilks96}
W.~R. Gilks and D.~J. Spiegelhalter, \emph{{Markov chain Monte Carlo in
  practice}}.\hskip 1em plus 0.5em minus 0.4em\relax Chapman \& Hall/CRC, 1996.

\bibitem{Asuncion07}
A.~Asuncion and D.~J. Newman, ``{UCI} machine learning repository,'' 2007.

\bibitem{FilipponeEA_MCMC_AIS13}
M.~Filippone, ``{B}ayesian inference for {G}aussian process classifiers with
  annealing and exact-approximate {MCMC},'' 2013, arXiv:1311.7320.

\bibitem{Neal01}
R.~M. Neal, ``{Annealed importance sampling},'' \emph{Statistics and
  Computing}, vol.~11, no.~2, pp. 125--139, Apr. 2001.

\bibitem{Ferri04}
C.~Ferri and J.~Hern{\'{a}}ndez-Orallo, ``{Cautious Classifiers},'' in
  \emph{ROCAI}, J.~Hern{\'{a}}ndez-Orallo, C.~Ferri, N.~Lachiche, and P.~A.
  Flach, Eds., 2004, pp. 27--36.

\bibitem{Moller98}
J.~M{\o}ller, A.~R. Syversveen, and R.~P. Waagepetersen, ``{Log Gaussian Cox
  Processes},'' \emph{Scandinavian Journal of Statistics}, vol.~25, no.~3, pp.
  451--482, 1998.

\bibitem{Chu05}
W.~Chu and Z.~Ghahramani, ``{Gaussian Processes for Ordinal Regression},''
  \emph{Journal of Machine Learning Research}, vol.~6, pp. 1019--1041, Dec.
  2005.

\end{thebibliography}
\end{document}